\documentclass[11pt]{article}

\usepackage[final]{acl}

\usepackage{times}
\usepackage{latexsym}

\usepackage[T1]{fontenc}
\usepackage[utf8]{inputenc}

\usepackage{microtype}

\usepackage{inconsolata}

\usepackage{graphicx}

\usepackage{hyperref}
\usepackage{url}
\usepackage{booktabs}
\usepackage{amsfonts}
\usepackage{amsmath}
\usepackage{amssymb}
\usepackage{nicefrac}
\usepackage{microtype}
\usepackage[table]{xcolor}
\usepackage{graphicx}
\usepackage{multirow}
\usepackage{cleveref}
\usepackage[most]{tcolorbox}
\usepackage{examples-slim}
\usepackage{placeins}

\newtcolorbox{controls}{
  colback=gray!5, colframe=gray!40, boxrule=0.4pt,
  left=6pt, right=6pt, top=4pt, bottom=4pt,
  fonttitle=\bfseries\small, title=Controls
}

\usepackage{tabularray}
\newcommand\cincludegraphics[2][]{\raisebox{-0.3\height}{\includegraphics[#1]{#2}}}
\usepackage{setspace}
\usepackage{subcaption}

\makeatletter
\DeclareRobustCommand*{\escapeus}[1]{%
    \begingroup\@activeus\scantokens{#1\endinput}\endgroup}
\begingroup\lccode`\~=`\_\relax
\lowercase{\endgroup\def\@activeus{\catcode`\_=\active \let~\_}}
\makeatother
\newcommand{\makesf}[1]{\textsf{{\escapeus{#1}}}}

\newcommand{\modelcomponent}{\mathcal{C}}
\newcommand{\sentbase}{\mathbf{b}}
\newcommand{\sentsrc}{\mathbf{s}}

\newcommand{\ptheta}{p_{\boldsymbol{\theta}}}

\newcommand{\gapbase}{\gap^{\sentbase}}
\newcommand{\gapsrc}{\gap^{\sentsrc}}
\newcommand{\gapmean}{\gap^{\modelcomponent \leftarrow \texttt{avg}}}
\newcommand{\gapintv}{\gap^{\modelcomponent \leftarrow \sentsrc}}

\newcommand{\amps}{\modelcomponent_\mathtt{A}}
\newcommand{\head}{\modelcomponent_\mathtt{H}}
\newcommand{\pros}{\modelcomponent_\mathtt{P}}
\newcommand{\antis}{\modelcomponent_\mathtt{N}}

\newcommand{\model}{\texttt{pythia-1b}}

\newcommand{\modelsmall}{\texttt{pythia-410m}}

\newcommand{\nheads}{8}              % heads per layer
\newcommand{\dmlp}{8192}             % MLP intermediate dimension (4 * dmodel)

\newcommand{\layerhead}[2]{\ensuremath{\text{L#1\,H#2}}}
\newcommand{\layerneuron}[2]{\ensuremath{\text{L#1\,n#2}}}

\newcommand{\pos}[1]{\text{pos}_{#1}}

\newcommand{\gap}{G}                     %  log p(correct) - log p(wrong)
\newcommand{\circuit}{\mathcal{C}_{\mathtt{full}}}

\usepackage[textsize=tiny]{todonotes}

\definecolor{takeawayblue}{RGB}{47,111,176}   % #2f6fb0, as in Figs. 2-3
\SetTracking{encoding=*, shape=sc}{70}
\newtcolorbox{takeaway}{
  enhanced, frame hidden, sharp corners,
  colback=white, boxrule=0pt,
  borderline north={0.5pt}{0pt}{black!70},
  borderline south={0.5pt}{0pt}{black!70},
  left=0pt, right=0pt, top=4.5pt, bottom=4.5pt, boxsep=0pt,
  before skip=9pt plus 2pt, after skip=9pt plus 2pt,
  fonttitle=\lsstyle\bfseries, coltitle=takeawayblue!75!black,
  attach title to upper={.\enspace},
  title=Takeaway
}

\title{The Emergence and Sparsification of a Syntactic Circuit}
\author{Qing Yao\textsuperscript{1} \quad Sasha Boguraev\textsuperscript{1} \quad Tiago Pimentel\textsuperscript{2} \quad Kyle Mahowald\textsuperscript{1}\\ \textsuperscript{1}The University of Texas at Austin \quad \textsuperscript{2}ETH Zurich \\ \texttt{\{qyao, sasha.boguraev, kyle\}@utexas.edu} \quad \texttt{tiago.pimentel@inf.ethz.ch}}

\begin{document}
\maketitle

\begin{abstract}
To understand the emergence of grammatical behavior in language models, we track circuits for subject--verb number agreement in \model{} and in ten seeds of \modelsmall{} across training.
We find that subject--verb agreement is learned early, and is initially handled by a diffuse collection of neurons.
However, after agreement \emph{behavior} has stabilized, the \emph{circuit} responsible for it continues to reorganize.
The attention heads which mediate agreement often shift during training: in the case of \model{}, behavior is controlled by two heads early in training but by a third (different) head later.
Furthermore, the circuit shrinks over training, with \model{}'s final circuit being strikingly sparse.
We take these results to be informative about how linguistic knowledge emerges and is represented in language models.
\end{abstract}

% ===========
% Github link
% ===========
\begin{tblr}{colspec = {Q[c,m]|X[l,m]}, stretch = 0}
    \cincludegraphics[width=1.2em, keepaspectratio]{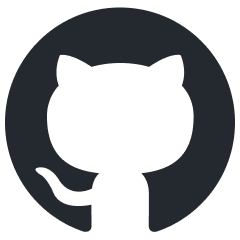}
     & \setstretch{.5}\href{https://github.com/dounick/sva-circuits}{{\makesf{sparsification-of-sva-circuits}}} \\
\end{tblr}

\section{Introduction}\label{sec:intro}

Our goal is to tell the story, in as much detail as possible, of how subject--verb number agreement emerges across training in language models (LMs).
Number agreement has long served as a probe for syntactic competence in language models \citep{linzen2016assessing, gulordava-etal-2018-colorless, marvin-linzen-2018-targeted, futrell-etal-2019-neural, mueller-etal-2020-cross, warstadt2020blimp,wei-etal-2021-frequency}.
%\tiago{I added gulordava-etal-2018-colorless here, since its multilingual, and Mueller as well. So dropped "English" from the sentence to make it more generally important.} 
It has also been used to probe how these models encode grammatical information internally \citep{finlaysonCausalAnalysisSyntactic2021, lasri-etal-2022-probing, arora-etal-2024-causalgym}.
Notably, this task's historical interest results from its symbolic specification, which, despite being simple, resists classic statistical methods.\footnote{E.g., it requires tracking structural information --- grammatical subjecthood --- over potentially unbounded distances, and can thus not be solved by simple Markov models.} 
A minimal circuit for handling it has to (1) recognize and store the grammatical number of the subject, (2) transport this information to the prediction position, and (3) license the corresponding output form.\looseness=-1

\begin{figure}
\centering
\includegraphics[width=\linewidth]{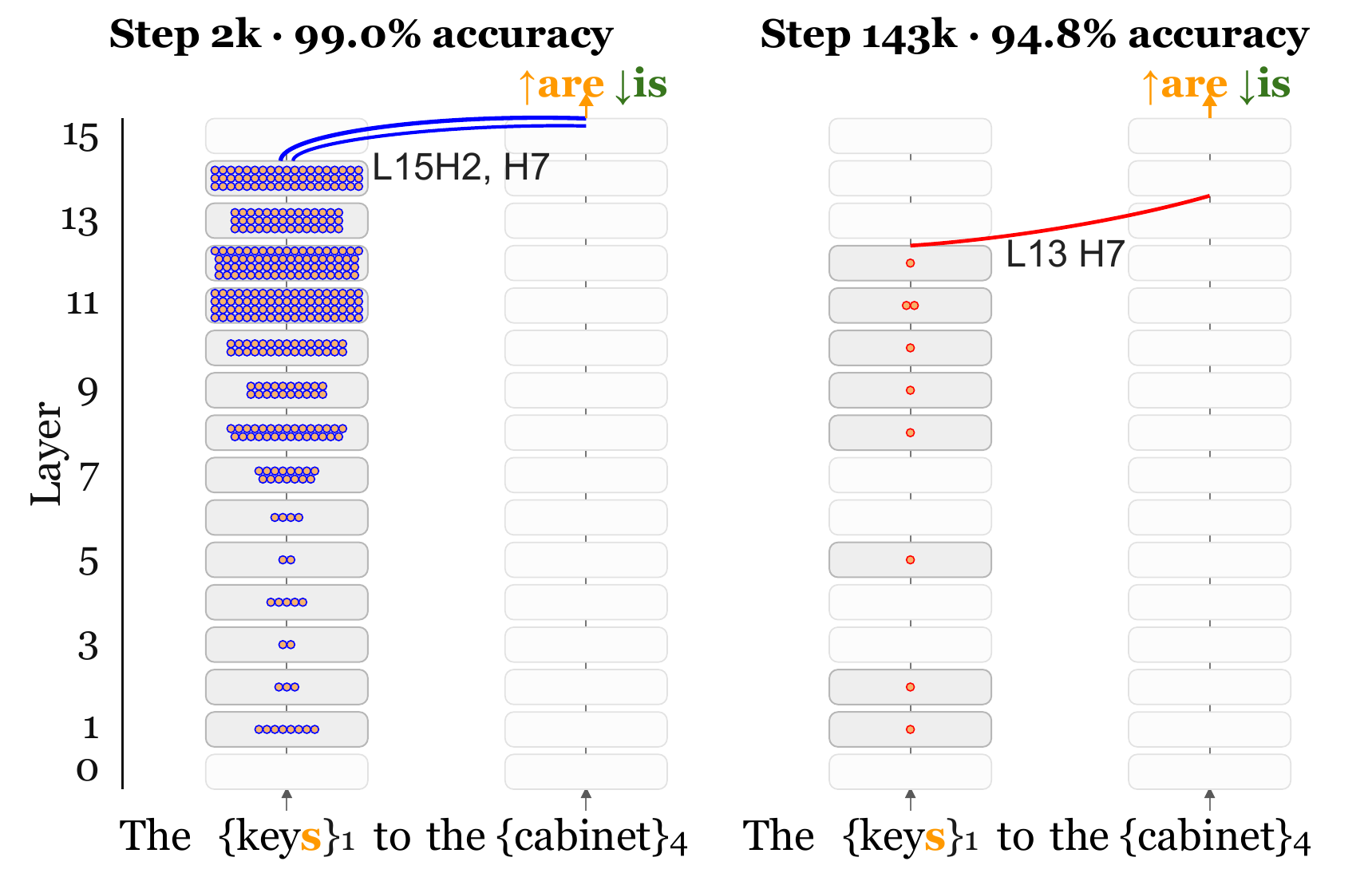}
\caption{The subject--verb agreement circuit for \model{} at step 2k (left) and step 143k (right).
While agreement is acquired early (within $2\%$ of training) and the model remains proficient thereafter, the underlying circuit continues to reorganize and sparsify.
Circles represent the smallest subset of top-ranked neurons whose sufficiency and necessity exceed 50\% (see \Cref{sec:behavior-vs-mechanism}).}
\label{fig:circuit}
\end{figure}

% \footnote{We take a broad definition of circuit here: any subset of a model's components or paths that connects a model's inputs to its outputs can be a circuit. E.g., a set of neurons compose a circuit when we  } 

In this work, we first (Experiment 1) trace the emergence of such a circuit over the course of language model training, studying ten seeds of \modelsmall{} and one seed of \model{}. 
We find that agreement is learned quickly and is initially controlled by a diffuse circuit.
Over the course of training, however, we find that while the model's performance remains mostly unchanged, the underlying circuit gradually evolves: it both sparsifies (concentrating in a smaller set of neurons) and is rerouted within the model (being implemented by different attention heads).
By the end of training, a surprisingly sparse circuit captures most of the agreement signal in \model{} (\Cref{fig:circuit}).
% Notably, this circuit transition happens gradually across training.

Such a gradual transition is consistent with recent reports of emergent syntactic behavior \citep{chen2024sudden}, and with work in developmental LM interpretability more broadly \citep{wangDifferentiationSpecializationAttention2024,chang-etal-2024-characterizing,fehlauer-etal-2025-convergence,kangaslahti2025hidden}, including recent work tracking the emergence of circuits longitudinally in open-checkpoint models \citep{tigges2024llm,ou-etal-2025-llms,hakimi-etal-2025-time,aljaafari2026emergence}.
This circuit's trajectory also parallels accounts of grokking \citep{power2022grokking,nanda2023progress,varma2023explaining,merrill2023a}, where a task is first learned by memorization, but that solution is replaced by a sparser and more general circuit after training accuracy has saturated. 
%Unlike grokking, we do not interpret our circuits' transition as one between memorization to generalization.

Later in our paper (Experiment 2), we zoom in on \model{}. 
We first track how the circuit evolves in this model, tracing how a subset of neurons remains relevant for agreement throughout training, but changes ``allegiance'' at some point, shifting the attention head they write to.
Second, we further sparsify this model's final circuit while preserving its performance.
We find that just 18 neurons and a single attention head can capture the primary channel for agreement in this model.
% Finally, we explore this circuits generalization to other template sentences
% for the primary agreement-mediating pathway at the final checkpoint, and analyze its properties.
% studying how \model{} reorganizes from a regime in which 2 particular attention heads control much of the behavior, to a regime controlled by a third (distinct) head. 
% and undertake a more bespoke analysis of the final state of its circuit, constructing the minimal full circuit for the final checkpoint.
We show this sparse circuit generalizes across different forms of subject--verb number agreement and to in-the-wild examples from WikiText \citep{merity2016pointer}.
Finally, we demonstrate that this circuit's information flows through a one-dimensional channel of the relevant transfer head. 
Overall, we show that mechanistic reorganization and sparsification continue beneath a stable agreement behavior.
In the case of \model{}, this culminates in a sparse and generalizable circuit for number agreement.\footnote{Throughout this paper, we focus on subject--verb number agreement, but often term it more compactly as `subject--verb agreement' or simply `agreement'.}
% \looseness=-1

\section{Finding agreement in LMs}

How can we understand and characterize the mechanisms responsible for a behavior in LMs?
The field of (mechanistic) interpretability focuses on answering this question.
Historically, researchers relied on diagnostic probes to find signals of the task within LMs' representations \citep[\textit{i.a.}]{alain2017understanding,hewitt-manning-2019-structural,hewitt-liang-2019-designing,pimentel-etal-2020-information,kulmizev-etal-2020-neural}.
In the years since, though, the field has grown into a substantial literature, with causality becoming one of its central aspects \citep{olah2020zoom,elhage2021mathematical,elazar2021amnesic}, and circuit-discovery
% \tiago{I changed this to "circuit-discovery" to unify the terminology (instead of using circuit-finding here instead)} 
a major thread within it \citep[][\textit{i.a.}]{wang2023ioi, Hanna2023GPT, conmy2023towards, stolfoMechanisticInterpretationArithmetic2023, hanna2024have}.

% \noindent
%\km{add initial sentence here, plus rest of Tiago comments}
% There is a strong precedence for using methods from interpretability to understand and characterize the internal representations recruited for agreement in LMs. 
% \citet{lakretz2019emergence} performed some of the first work on this phenomenon, identifying a circuit of roughly this shape in small LSTMs. While this analysis was illuminating, it was bespoke to both the phenomenon and model studied. 
% However, in the years since, mechanistic interpretability has grown into a substantial literature, with circuit-finding\tiago{should we unify this term as "circuit-discovery"} becoming a major thread \citep[][\textit{i.a.}]{wang2023ioi, Hanna2023GPT, conmy2023towards, stolfoMechanisticInterpretationArithmetic2023, hanna2024have}.
% 
% Early circuit-discovery work primarily focused on finding subgraphs of transformer attention heads responsible for a given task. These methods provided notable insights regarding the agreement mechanisms in LMs, including the presence of syntax-aware heads in BERT \citep{clarkWhatDoesBERT2019} and an analagous language-agnostic number-transport head and downstream readout neurons for English and Spanish processing in \texttt{gemma-2-2b} \citep{ferrando2024similarity}.

Notably, some of the earliest work on this topic tried to identify the mechanisms behind number agreement in LMs.
Studying LSTMs, \citet{giulianelli-etal-2018-hood} and \citet{lakretz2019emergence} identified components or circuits in these models which implemented number agreement.
Studying transformer models, \citet{clarkWhatDoesBERT2019} identified syntax-aware attention heads in BERT, and \citet{ferrando2024similarity} identified a bilingual number-transport attention head for English and Spanish in \texttt{gemma-2b}.
% \tiago{I changed this to Gemma 2b, since I think that's the correct model. qing: the paper authors called it gemma 2b but the huggingface name is gemma-2-2b, we'll just go with the old version}
% While this analysis was illuminating, it was bespoke to both the phenomenon and model studied. 
% 
More recently, there has been a shift towards trying to identify more fine-grained circuits within LMs.
\citet{marks2025sparse}, for instance, used sparse autoencoders (SAEs) to identify sparse circuits (with fewer than 250 nodes)
responsible for subject--verb agreement in both \texttt{pythia-70m} and \texttt{gemma-2-2b}.
Building on such work, \citet{arora2026languagemodelcircuitssparse} showed that circuits over MLP activations can be as sparse as SAE-based ones.
% in with subject--verb agreement as one of their test-beds. 
% Together, these works suggest that it is both possible and fruitful to study LMs by searching for sparse circuits responsible for specific behaviors---and, in particular, for how they implement subject--verb number agreement.

Most relevantly, \citet{tigges2024llm} trace Pythia circuits across four tasks, one of which is subject--verb agreement.
They find that the underlying algorithm implemented by the circuit remains stable across training, but that the circuit's components and size can change.
While they focus on a variety of tasks and localize their circuit at a coarse level of MLP blocks and attention heads, we zoom in on a single task, studying its behavior across many random seeds and focusing our analyses at the neuron level.
We track how many and which neurons are needed to causally control agreement across training, we characterize how they reroute between number-transport heads, and we describe a complete subject-to-prediction circuit in the model.\looseness=-1

Overall, our work provides three primary contributions.
First, building on prior longitudinal circuit analyses across Pythia checkpoints and scales \citep[e.g.,][]{tigges2024llm}, we study subject–verb agreement at finer granularity, combining head- and neuron-level interventions to track the agreement circuit across checkpoints and random seeds. 
Second, we characterize not only which components change in this circuit, but how agreement-relevant neurons sparsify and reroute as the dominant transport head changes.
Finally, we show that most of the agreement behavior in \texttt{pythia-1b} can be captured by a sparse circuit.

\section{Methods and materials}
% \sasha[]{high level things I think we might have lost in structure and polishing: how many examples do we find the circuit and evaluate on, discuss that these are disjoint sets,}
\label{sec:model-and-task}

\subsection{Models}
We study models of the \texttt{pythia} family \citep{biderman2023pythia}. 
We focus our analysis on $\model{}$ --- a 16-layer autoregressive transformer with $\nheads$ attention heads per layer, a hidden size of 2048 neurons, and MLPs with intermediate dimension $\dmlp$ --- and on the ten seeds of \modelsmall{}, nine of which are from PolyPythias \citep{vanderwal2025PolyPythias}, each with 24 layers, 16 attention heads per layer, hidden size 1024, and MLP dimension $4096$. 
By performing our analysis across model sizes and seeds, we study the behavior and internal representations of the models more generally.

\subsection{Subject--verb agreement task} \label{sec:data}

We study subject–verb agreement, focusing on sentences with an intervening prepositional phrase, a task we refer to as \textsc{PP} for convenience. 
An exemplar minimal pair can be seen in (1).
\begin{examples}
    \item \begin{examples}
        \item[\textbf{b}] The$_0$ \textbf{key}$_1$ to$_2$ the$_3$ cabinet$_4$ $\rightarrow$ \textbf{is} 
        \item[\textbf{s}] The$_0$ \textbf{keys}$_1$ to$_2$ the$_3$ cabinet$_4$ $\rightarrow$ \textbf{are}
    \end{examples}
\end{examples}
We denote token roles with subscripts: $\pos{1}$ for the subject position, $\pos{2}$ for the last token position of the preposition, and $\pos{4}$ for the final token/prediction position.
In order to generate a large number of unique minimal pairs, we adapt the templates from \citet{arora-etal-2024-causalgym}.
Specifically, we restrict the lexical inventory to single-token nouns and verbs (in both singular and plural forms) and only generate minimal pairs differing in subject number. 

In \Cref{sec:circuit}, we evaluate how the circuits we discover generalize to other agreement examples.
Beyond PP sentences, in that section, we also investigate sentences with a subject relative clause distractor (\textsc{SubjectRC}), an object relative clause distractor (\textsc{ObjectRC}), or no distractor (\textsc{Simple}).
% We generate these in the same way as above.
% , and introduce a simple template

For the \textsc{PP} task, we create a training set of 400 minimal pairs, balanced in the number of subjects and distractors.
We use this set to identify candidate circuit components.
We also create disjoint evaluation sets for \textsc{PP} and the other templates, each consisting of 192 balanced minimal pairs (with the exception of \textsc{Simple}, which has 32 minimal pairs). For both behavioral validation (\Cref{sec:method-behavior}) and circuit evaluation (\Cref{sec:circuit-evaluation}), we use the evaluation sets.
For more details on all templates, along with associated exemplars, see App.~\ref{app:templates}.

\subsection{Behavioral validation}\label{sec:method-behavior}

To measure the behavioral performance of the studied models on subject--verb agreement, we calculate their accuracy on the evaluation set using the following paradigm. A model is deemed to correctly process a minimal pair if for each item, the probability for the correct continuation is higher than the alternative.
That is, given the minimal-pair sentences in (1), a model is accurate if it correctly predicts for \emph{both} sentences that
\begin{subequations}
\label{eq:example_probs}
\begin{align}
    &p(\text{is}|\text{The key ...}) > p(\text{are}|\text{The key ...}) \, \texttt{ and } \\ 
    &p(\text{is}|\text{The keys ...}) < p(\text{are}|\text{The keys ...}).
\end{align}
\end{subequations}
% $p(\text{is}|\text{The key}) > (\text{are}|\text{The key})$ and that $p(\text{is}|\text{The keys}) < (\text{are}|\text{The keys})$. 

\subsection{Circuit evaluation}\label{sec:circuit-evaluation}

Let $\modelcomponent$ be a model component, or a collection of components, proposed to be a part of a circuit of interest and thus responsible for a certain task.
To evaluate the causal effect of $\modelcomponent$ on that task, we intervene on its activations to calculate its \textbf{necessity} and \textbf{sufficiency} for downstream behavioral control. This section details these metrics.

% \paragraph{Mean and exchange interventions.}

% \paragraph{Agreement Gap.} 
Both of these metrics rely on a metric we refer to as the \textbf{agreement gap}, $\gap$.
Let $\sentbase$ and $\sentsrc$ be base and source minimal-pair sentences, with $\ell_\sentbase$ and $\ell_\sentsrc$ their respective continuations.
In ex.~(1), for instance, $\ell_\sentbase = \textit{is}$ and $\ell_\sentsrc = \textit{are}$.
Given this pair, $\gap$ is the expected difference in log probability between the base and source continuations:
% \sasha[]{introduce $\ell$ \response{tiago} done!}
% \tiago{unclear to me what is the base, and what's the source. Should the model agree with base or source?}
% 
\begin{equation}\label{eq:agr-gap}
\begin{aligned}
    \gapbase &= \underset{(\mathbf{b}, \mathbf{s}) \in D}{\mathbb{E}} \log\left(\frac{ \ptheta(\ell_\mathbf{b} | \sentbase)}{ \ptheta(\ell_\mathbf{s} | \sentbase)}\right), \\
    \gapsrc &= \underset{(\mathbf{b}, \mathbf{s}) \in D}{\mathbb{E}} \log\left(\frac{ \ptheta(\ell_\mathbf{b} | \sentsrc)}{ \ptheta(\ell_\mathbf{s} | \sentsrc)}\right)
\end{aligned}
\end{equation} 
where $\ptheta$ stands for the evaluated model run under a specific configuration. 
$\gapbase$ and $\gapsrc$ therefore represent agreement gaps of the model when processing the base sentence or source sentence, respectively. A higher preference for the base continuation corresponds to a more positive gap score. For a behaviorally correct model, $\gap^\sentbase$ should be positive, while $\gap^\sentsrc$ should be negative.
% the models agreement gap when the model processes the base input 
% better performance on the task.
% 
% \noindent To make this metric concrete, consider the computation for a single base sentence \textit{``The key \{is/are\}''} and a source sentence \textit{``The keys \{are/is\}''}. \gap{} for the base sentence would be the log-probability difference between the correct completion (``is'') and the incorrect one (``are''). 
Our notions of sufficiency and necessity then measure how much this gap changes under manipulations of activations.

% \paragraph{Necessity.} 
A model component $\modelcomponent$ is \textbf{necessary} if performing a mean ablation \citep{wang2023ioi} causes the model's agreement gap to drop to near-zero:
% We define necessity as:
\begin{equation}\label{eq:necessity}
    \operatorname{Nec}(\modelcomponent) = \frac{\gapbase - \gapmean}{\gapbase}
\end{equation}
where $\gapmean$ is the agreement gap on input $\sentbase$ when we replace $\modelcomponent$'s activations with their average value over a training set.\footnote{Mean activations are computed on training sets for the \textsc{PP} template and on the evaluation set for other templates.}

% Activation means are frozen and taken from the training sets, but our necessity numbers are computed over the evaluation sets.\sasha[]{appendix?}

% \paragraph{Sufficiency.} 
A model component $\modelcomponent$ is \textbf{sufficient} if exchanging its activations across minimal pairs $\mathbf{b} \leftarrow \mathbf{s}$ shifts the verb predictions for $\mathbf{b}$ towards those of $\mathbf{s}$. This is measured by the following metric:
\begin{equation}\label{eq:sufficiency}
    \operatorname{Suff}(\modelcomponent) = \frac{\gapbase - \gapintv}{\gapbase - \gapsrc},
\end{equation} 
\noindent where $\gapintv$ is the agreement gap when the model is run with input $\sentbase$ with $\modelcomponent$'s activations replaced by their values under input $\sentsrc$.
Intuitively, a component is sufficient if intervening on $\modelcomponent$'s activations has roughly the same causal effect as directly swapping the subject token for its singular/plural counterpart.\looseness=-1
% % To make this concrete, consider the computation for a single base sentence ``The key \{is/are\}'' and a source sentence ``The keys \{are/is\}''. 
% % \gap{} for the base sentence would be the difference between the correct completion (``is'') and the incorrect one (``are''). 
% $\gapintv$ would measure the agreement gap after the intervention of having surgically inserted ``keys' material into the ``key'' sentence.
% $\gapbase  - \gapsrc$ is in effect then a measure of how big the agreement gap would be if the intervention were ``perfect'' at turning the base sentence into the source sentence. 
% % In our toy example, we obtain this empirically by looking at the gap between ``The keys is'' and ``The keys are'' (which we expect to be negative) to get $\gap_\text{target}$, which we then subtract from $\gap$ (where $\gap${} is the ``The key is'' vs. ``The key are'' difference).

% $\mathcal{C}$ is \textbf{sufficient} for agreement if intervening on $\mathcal{C}$ is comparable to swapping out the subject number directly, i.e. $\text{range}_\mathcal{C} \approx 1$.\tiago{Does range measure sufficiency then? Figures 2 and 3 imply sufficiency is only the G intv, not range.}

\paragraph{Relation to other metrics.} Existing circuit work evaluates \textbf{completeness} and \textbf{faithfulness} \citep{wang2023ioi, marks2025sparse, arora2026languagemodelcircuitssparse}.
Within a predefined scope (global set of relevant components), a circuit is \textbf{complete} if ablating the circuit degrades performance as much as ablating the entire scope, and \textbf{faithful} if ablating its complement preserves performance. 
Our notion of necessity is conceptually similar to completeness, as ablating the entire scope should typically bring the model close to chance performance --- and thus near zero agreement gap.
We choose to evaluate sufficiency rather than faithfulness as defining a circuit's complement (to ablate) is difficult, especially when evaluating single components.

\subsection{Circuit discovery methods}\label{sec:methods}

This section details the methods we use to localize our circuit.
Throughout our experiments, we discover and evaluate our circuits on the \textsc{PP} template; we later show that the discovered circuit generalizes to the other forms of number agreement.

\paragraph{Head identification.} We identify attention heads responsible for transporting a subject's number by measuring each head's necessity and sufficiency at the prediction position, $\pos{4}$, across the training set. We intervene on the full head to compute these metrics, not only on its subject--prediction values.\footnote{After a head is selected as a circuit component, it is then evaluated for necessity and sufficiency on the evaluation sets.\looseness=-1}
% \looseness=-1

\paragraph{Neuron identification.}
We follow \citet{arora2026languagemodelcircuitssparse} in using Relevance Patching \citep[RelP;][]{jafari2025relp}, a gradient-attribution method, to localize neurons responsible for LM behavior. We utilize the unpaired variant of RelP which scores components by the product of their activation on the base input and the gradient of a task metric with respect to that component. It differs from attribution patching methods \citep{nanda2023attribution, kramar2024atp, syed2024attribution} by computing the gradient on a locally linear replacement model --- where in the backward pass each nonlinearity is frozen at its forward value. This results in a less noisy and more faithful gradient \citep{jafari2025relp}. Details of our RelP implementation, including the linearization rules, an integrated-gradient robustness check, and task-specific modifications, are in App.~\ref{app:attribution-details}.

We score each MLP neuron, $\modelcomponent_n$, as in \eqref{eq:relp}, where $v_n(\mathbf{b})$ is $\modelcomponent_n$'s activation on the base input, $m$ is the per-example base agreement gap, and the gradient is taken relative to the replacement model:
% \begin{equation}
\begin{align}\label{eq:relp}
    &\mathrm{RelP}(\modelcomponent_n) =
  \underset{(\mathbf{b}, \mathbf{s}) \in D}{\mathbb{E}}\!\left[\,
    v_n(\mathbf{b})\,
    \frac{\partial m(\mathbf{b}, \mathbf{s})}{\partial v_n(\mathbf{b})}
  \,\right], \\
  &\qquad\qquad \texttt{where } m(\mathbf{b}, \mathbf{s}) = \log\frac{\ptheta(\ell_{\mathbf b}\mid\sentbase)}
            {\ptheta(\ell_{\mathbf s}\mid\sentbase)}. \nonumber
\end{align}
% \end{equation}
A positive RelP score indicates that $\modelcomponent_n$ pushes the model toward verbs licensed by their subject.
We apply this method to score every post-GELU MLP neuron in each layer at the relevant token positions for a given analysis. As RelP is an attribution method, we treat it as a hypothesis generator for causally-relevant neurons. To ensure their importance, we then causally validate such candidate neurons through interventions detailed in \Cref{sec:circuit-evaluation}.

% , and as such should be only considered as a candidate generation method. Depending on the experiment, we may rank neurons by signed or absolute valued RelP scores, and pool candidates across positions.\footnote{We note that RelP scores are not directly comparable across positions. We treat the pooled neurons as a candidate pool that is sufficiently large to include the most relevant neurons from each position. See App.~\ref{app:pooling} for a discussion showing that this does encompass all relevant neurons across positions.} We state these choices separately for each experiment. In all cases, we causally validate components selected from the ranked candidate lists by computing necessity or sufficiency. Details of our RelP implementation, including the linearization rules, an integrated-gradient robustness check, and task-specific modifications, are in App.~\ref{app:attribution-details}.

\paragraph{Neuron and head routing.}
Having separately identified neurons and heads relevant to agreement, we want to establish routing relations between them. For instance, given two heads that could both transport number information, we want to determine which route a candidate neuron takes.
To this end, we attribute each neuron to heads (possibly to both or neither) using a path-patching-like method \citep{wang2023ioi,goldowskydill2023localizingmodelbehaviorpath}. For neurons before the prediction position, we (1) intervene on the neuron and measure the downstream causal effect and (2) intervene on the same neuron while holding a head $\modelcomponent_h$'s corresponding value edge fixed at the base activation. Any effect that drops out under the second intervention relative to the first is attributed to the route via $\modelcomponent_h$. Prediction-side neurons are likewise classified by which head interventions they read out from. We describe our methodology in detail in App.~\ref{app:routing-attribution-details}.

\section{Experiment 1: Behavior--mechanism decoupling in training}\label{sec:emergence}
In our first experiment, we track the internal organization of \model{} and the ten seeds of \modelsmall{} as they develop the capacity to perform subject--verb agreement. 
% We follow the methodology presented in \Cref{sec:methods}. 
At each checkpoint, we measure their agreement capabilities, identify the neurons associated with agreement using RelP, and determine which heads are sufficient and necessary for agreement. Our goal here is not yet to produce complete agreement circuits, but to provide a description of the models' behavior and identify agreement-relevant components through training. 
% Below, we detail our results, focusing on the PP template.

\subsection{Agreement behavior is learned within 2\% of total training steps}
First, we measure the behavioral performance of our studied models across training. 
The y-axis in \Cref{fig:behavior-vs-mechanism} shows our results. 
At steps 256 and 512, the models have not learned agreement yet, always preferring singular continuations over plural ones, regardless of subject number. 
Since agreement accuracy requires predicting both singular and plural verbs correctly, through step 512, no model is even $2\%$ accurate on agreement.
By step 2k, though, agreement is high for all ten \modelsmall{} seeds, with a median accuracy of $96.9\%$ (range $88.5$--$100\%$), and a median agreement gap of $3.35$. At the end of training, the models' median accuracy and gap are $91.7\%$ and $3.39$, respectively. This shows agreement capabilities are learned early in training and remain relatively stable throughout.
The larger \model{} shows the same pattern --- its accuracy and gap change from $99.0\%$ and $3.21$ at step 2k to $94.8\%$ and $3.48$ at the final checkpoint.\footnote{The better performance at step 2k compared to the final checkpoint is somewhat mysterious. The final checkpoint errors are cases with opposite-number distractors, suggesting that models are perhaps falling for agreement attraction effects. Broadly, though, this result is consistent with the ``unruly'' BLiMP curves shown in \citet{bunzeck-zarriess-2024-fifty}.}

Overall, our results are consistent with \citet{vanderwal2025PolyPythias}'s finding that linguistic knowledge is largely acquired by $10^3 - 10^4$ steps. 
This raises a question, though: once agreement behavior is saturated, is the mechanism behind this syntactic competence effectively learned and frozen in place, or does it continue to evolve in ways that behavioral measurements do not reveal? 
To answer this question, we first track the neurons that process subject number at the subject, and then the attention heads that transport it to the prediction position.

\begin{takeaway}
The models learn agreement after only around 2k training steps, with behavior remaining generally consistent thereafter.
\end{takeaway}

\subsection{Circuit neurons sparsify through training}\label{sec:behavior-vs-mechanism}
We now consider the neurons implicated in the agreement circuits at the subject position, and compare how they change across training checkpoints. 
In particular, if the agreement circuit is effectively ``finalized'' after proficiency is achieved in this task, the number of circuit neurons should remain stable.

For each model and checkpoint, we identify the neurons most responsible for agreement by (1) computing a RelP score for all MLP neurons across all layers at the subject position on the training set, (2) ranking them in descending order, since a larger positive RelP score indicates stronger support for agreement, and (3) evaluating their causal effect on the model.
For (3), we define $k_{50}$ as the smallest $k$ such that a circuit consisting of the top-$k$ ranked neurons, $\modelcomponent_k$, is at least $50\%$ necessary and sufficient under intervention:\footnote{Finding the precise value of $k$ is prohibitively expensive. In practice, we compute the score in \eqref{eq:k50} at intervals of powers of two, and interpolate linearly in $\log_2$ space for an estimate.}
% 
% \vspace{-10pt}
\begin{equation}\label{eq:k50}
    \resizebox{.85\columnwidth}{!}{$\displaystyle
        k_{50} = \min\left\{ k \mid \min(\operatorname{Nec}(\modelcomponent_k), \operatorname{Suff}(\modelcomponent_k)) \geq 50\% \right\}.
    $}
\end{equation}

% The y-axis is agreement accuracy. 
% The figure makes clear number-controlling neurons become increasingly concentrated through training for \model{} and for 8 of 10 seeds of \modelsmall{}, even as agreement accuracy is kept constant. \qy{add results for different k thresholds in appendix}
\begin{figure}
    \centering
    \includegraphics[width=1\linewidth]{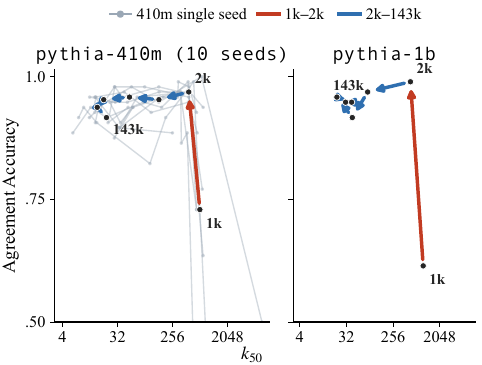}
    \vspace{-20pt}
    \caption{Agreement accuracy vs. \# of subject-neurons in the circuit through training across our models. Each point is a checkpoint. Typically, steps 1k--2k form the agreement acquisition phase. In the remaining steps, the circuit sparsifies before reaching some equilibrium.}
    \label{fig:behavior-vs-mechanism}
    \vspace{-5pt}
\end{figure}

Our results are in \Cref{fig:behavior-vs-mechanism}.
We find that the agreement circuit tends to sparsify throughout early training --- even as performance remains constant --- leveling off around halfway through training.
Further, the same neurons are generally deemed important across checkpoints.
For example, in \model{}, we find that all 21 neurons in step 18k's $k_{50}$ are present in the larger 42-neuron $k_{50}$ at step 8k, with 20 also appearing in the 41-neuron $k_{50}$ at step 50k, and 18 remaining in the final checkpoint's $k_{50}$.
The same pattern holds across most runs of \modelsmall{}: taking the intersection of the $k_{50}$ neuron sets of all post-8k checkpoints, we find a median of $85\%$ overlap with each seed's smallest $k_{50}$ neuron set. 
For more details on neuron overlap across checkpoints and on whether results hold across thresholds of $k$, see App.~\ref{app:exp1-details}.

Notably, we observe a substantial increase in $k_{50}$ during training for two \modelsmall{} seeds (3 and 4).
Interestingly, these seeds are exactly the training outliers identified by \citet{vanderwal2025PolyPythias}, which underperformed on benchmarks after suffering loss spikes during training.
% that caused these models to regress to earlier parameter states.
These loss spikes occur precisely where we see sharp rebounds in individual $k_{50}$ counts (at steps 83k and 123k for seeds 3 and 4, respectively; more details in App.~\ref{app:outlier}).

This analysis tells us that agreement circuits become sparser at the subject position during training, involving fewer neurons.
But the complete circuit also involves transporting that signal to the prediction position. 
Next, we ask whether the number-transporting heads remain stable after agreement is learned, or if they also undergo reorganization.

\begin{takeaway}
After models learn agreement, their circuits still sparsify, with progressively fewer neurons being required to capture the same level of causal control over agreement across training.
\end{takeaway}

\subsection{Number is carried by a small set of heads}
\begin{figure}[t]
      \centering

      \begin{subfigure}{0.98\linewidth}
          \centering
          \includegraphics[width=\linewidth]{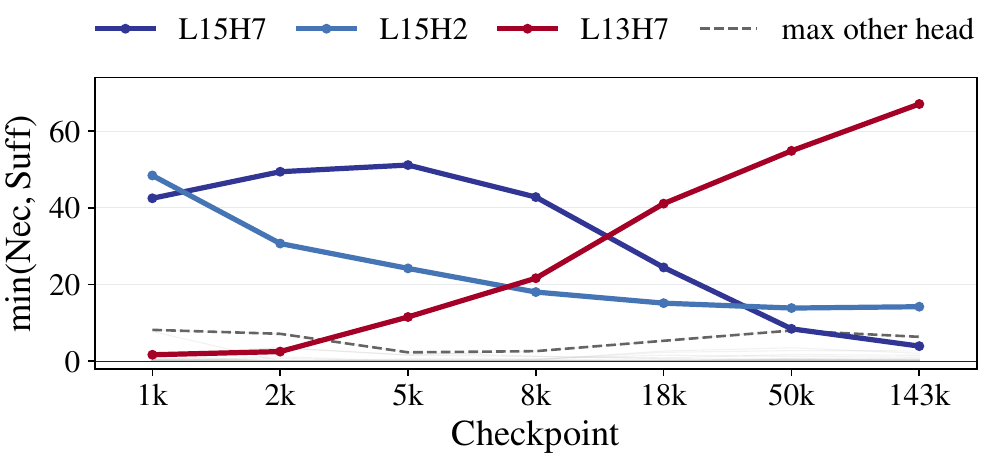}
          \caption{Head scores across training for \model{}.}
          \label{fig:head-scan}
      \end{subfigure}

      \vspace{0.5em}
      
      \begin{subfigure}{0.98\linewidth}
          \centering
          \includegraphics[width=\linewidth]{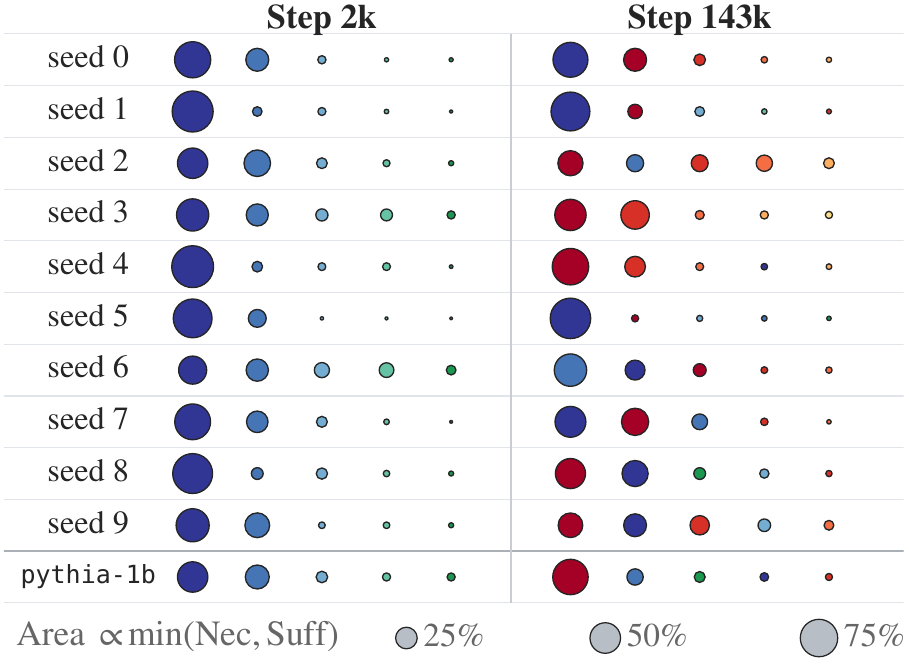}
          \caption{Top-five heads across model runs.}
          \label{fig:head-top5}
      \end{subfigure}
      \caption{Attention-head reorganization. \textbf{Top:} In \model{}, attention heads reorganize across training checkpoints, with the heads important early, \layerhead{15}{7} and \layerhead{15}{2}, ceding importance to \layerhead{13}{7}. All other heads are negligible. \textbf{Bottom:} The identities of the leading heads change across training. Circle area is proportional to each head's individual $\min\{\operatorname{Nec},\operatorname{Suff}\}$ score, with each individual head a different color.}
    \label{fig:heads-across-training}
\end{figure}

We now identify the attention heads in the agreement circuits at the prediction position by similarly measuring necessity and sufficiency across training. \Cref{fig:head-scan} demonstrates that in \model{}, early agreement circuits are almost entirely mediated by two heads ($\layerhead{15}{7}$ and $\layerhead{15}{2}$). However, by step 18k, head $\layerhead{13}{7}$ becomes the most important, and remains so through the final checkpoint (although $\layerhead{15}{2}$ persists as a weaker secondary channel). In comparison, all other heads contribute negligibly.

\Cref{fig:head-top5} tells a similar story for \modelsmall{}. Across all models, a small set of heads is responsible for carrying agreement information, often with a clear leading head individually achieving over $50\%$ in necessity and sufficiency. Despite consistently small numbers of relevant heads, none of the examined models have the same top-3 heads at steps 2k and 143k. (Detailed results in App.~\ref{app:head-across-seeds}.)
% For instance, for \model{} at step 2k, agreement is almost entirely mediated by two heads ($\layerhead{15}{7}$ and $\layerhead{15}{2}$), with the eventual winner ($\layerhead{13}{7}$) being dormant at that stage (as seen in \Cref{fig:head-scan}).
% By step 18k, head $\layerhead{13}{7}$ becomes the most important, and remains so through the final checkpoint (although $\layerhead{15}{2}$ persists as a weaker secondary channel).

\begin{takeaway}
We find that agreement is generally mediated through a small set of late-layer attention heads, and the leading transport heads often shift even after behavior stabilizes.
\end{takeaway}

\begin{figure*}
    \centering
    \includegraphics[width=1\linewidth]{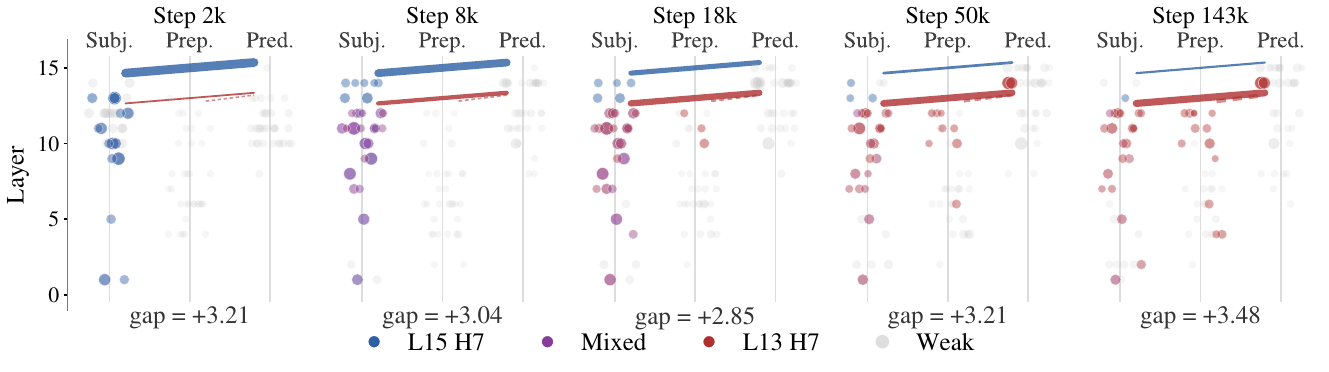}
    \vspace{-25pt}
    \caption{Top-32 ranked RelP neurons across positions and attention heads $\layerhead{13}{7}$ and $\layerhead{15}{7}$ across checkpoints as identified on \textsc{PP} training set ($N=400$). Neurons are colored by their correspondence to the respective heads, on a gradient between blue ($\layerhead{15}{7}$) and red ($\layerhead{13}{7}$), determined by path patching. Gray neurons are not attributed to either route. Line width indicates each head's sufficiency. We see that throughout training, there is a gradual transition from an $\layerhead{15}{7}$-routed mechanism to an $\layerhead{13}{7}$-routed one.}
        \label{fig:circuit-emergence}
\end{figure*}

\section{Experiment 2: The sparsification of the \model{} circuit}\label{sec:circuit}
% \tiago{Minor nitpick: Capitalisation in section headers is not 100\% consistent. And we could add a Pythia-1b macro for plaecs where it needs to be capitalised? Or make the section titles non title case by default\response{qing: i'm going to make them all capitalize first word, and first word after colon, how does that sound?}}
Having shown the general developmental trajectory of agreement circuits across seeds and models, we now zoom in on the final circuit for one model: \model{}. First, we show that its neuron- and head-level reorganizations happen in tandem. Next, we prune the circuit down to 18 neurons and one attention head. We then demonstrate this hand-constructed sparse circuit generalizes to novel templates and in-the-wild text. Finally, we show the head's transport can be further localized to just a one-dimensional subspace of its output channel.

% In this section, we focus on zooming in on the behavior of one seed from one model, to understand what is happening when \model{} reorganizes from a 2-head mechanism to a 1-head mechanism.
% Our goal is first to understand the re-organization that takes place and, second, to then analyze the final circuit in detail.

\subsection{Joint reorganization}\label{sec:circuit-comparison}

In \Cref{sec:emergence}, we showed that while agreement is learned early, its circuit reorganizes during training. Two changes occur together, namely that the number of subject-neurons in the circuit decreases, and the heads transporting the number information shift from a L15 route to a \layerhead{13}{7} route. In this section, we link these observations and show that circuit neurons change the pathway they feed into or read out from as the dominant head changes.

To this end, we first collect the top-32 ranked neurons according to RelP at each checkpoint across three token positions: subject $\pos{1}$, preposition $\pos{2}$, and prediction $\pos{4}$. We then apply the routing attribution method from \Cref{sec:methods} to measure the extent to which a neuron reads from or writes to \layerhead{15}{7} (the head primarily responsible for information transport early in training) or \layerhead{13}{7} (the head responsible late in training).

\Cref{fig:circuit-emergence} shows our results across checkpoints.
We see that as the relative head importance changes, the neurons follow. That is, early in training, we see neurons route through \layerhead{15}{7}, the early primary transport channel. However, as training progresses and \layerhead{13}{7}'s channel emerges, pre-L13 neurons begin to show mixed routing. Then, as \layerhead{13}{7} becomes increasingly prominent, most neurons reorganize to start writing to or reading from that channel.
By the final checkpoint, both upstream and downstream neurons are primarily aligned with the \layerhead{13}{7} channel.

\begin{takeaway}
    The agreement circuit in \model{} shifts from a route mediated predominantly by \layerhead{15}{7} to one mediated by \layerhead{13}{7} throughout training. As the heads shift, the neurons follow.
\end{takeaway}

\subsection{A final sparse circuit}

Having shown the trajectory of the joint neuron and head circuit through training, we now zoom in on the final checkpoint of \model{} to produce a sparse circuit for agreement.

\paragraph{Circuit construction.}
Our goal here is to construct a sparse circuit for subject--verb agreement. 
To do so, we made a series of choices that we believe would represent the model's internal agreement mechanism while minimizing the circuit size.

We started with two qualitative observations based on the results of the earlier analyses and further manual inspection of the results. 
First, number agreement at the final checkpoint is largely carried by one head, $\layerhead{13}{7}$. 
Second, MLPs propagate agreement information across layers, while self-attention is not strongly involved in this process.\footnote{We show in App.~\ref{app:scope} that ablating self-attention has negligible effect on agreement.}

Our goal was then to construct a circuit consisting of the one relevant head and a set of relevant neurons that captures the role of the MLPs in processing agreement.
To this end, we again used RelP to identify a set of potentially important neurons for the circuit. 
We further augmented this set with a few extra hand-picked neurons, identified as important via other methodologies.\footnote{We performed an initial analysis, prior to applying RelP, using Distributed Alignment Search \citep[DAS;][]{geiger2024DAS}, a supervised interpretability method which localizes LM behavior to causally sufficient subspaces in the higher-dimensional activation space. Consistent with previous works \citep[e.g.][]{arora-etal-2024-causalgym}, we found that the number feature could be localized to a single dimension of the activation space, but this subspace appeared to change across layers due to MLPs writing additional number information into the residual stream. This motivated us to identify important neurons which write to these DAS subspaces. For each neuron, we computed its number-dependent contribution to the next-layer as the product of its paired activation difference between singular and plural inputs and the neuron's output vector. The cosine similarity of the resulting contribution with the next-layer DAS subspace was then used to rank neurons.
% This motivated us to ask whether a few representative neurons could capture the roles of these MLPs.
% 
DAS settings and the neuron selection and pruning procedures are in App.~\ref{app:discovery}.}
This resulted in a small set of 21 candidate neurons.
% 
% 
% Each neuron writes along a fixed output vector into the residual stream (the corresponding column of the down-projection matrix). For each neuron, we computed its paired activation difference between singular and plural inputs averaged across the training set, and multiplied this difference by the neuron's output vector. The cosine similarity of the resulting number-dependent contribution with the next-layer DAS subspace is the neuron's score. 
% 
% This gave us a small set of neuron candidates, from which we constructed an initial circuit of one head and 16 neurons. We later applied RelP, which ranked many of these neurons highly, but also generated five candidates we originally missed. We added these candidates, then 
We then greedily pruned neurons whose removal had negligible effects on circuit necessity and sufficiency. This procedure removed three redundant neurons, leaving us with our final circuit. 

Notably, this circuit's construction process involved several judgment calls. 
Our goal was not to provide a systematic solution for its creation, but to try to qualitatively understand and improve the circuit iteratively.
There are no guarantees that this circuit is maximally performant or sparse.
At the end of the procedure, we had a sparse circuit, $\circuit$, which contains 18 neurons and one attention head.
Ten neurons are at the subject position $\pos{1}$ and upstream of the head \layerhead{13}{7} (in L1--L12 MLPs); eight neurons are at the prediction position $\pos{4}$ and downstream of the head (in L14--L15 MLPs).

\begin{table}[t]
\centering
\small
\begin{tabular}{l c c}
\toprule
\textbf{Components} $\modelcomponent$                                & $\mathbf{\gap}$ & \textbf{Measure}\\
\midrule
\rowcolor{gray!30} \textit{\textbf{Necessity}} & $\mathbf{\gapmean}$ & \textbf{Nec.} \\
$\circuit$  & $+0.76$ & $78\%$  \\
$\modelcomponent_\text{scope}$               & $+0.58$ & $83\%$  \\ \midrule
\rowcolor{gray!30} \textit{\textbf{Sufficiency}} & $\mathbf{\gapintv} $ & \textbf{Suff.} \\
$\circuit$  & $-2.29$ & $83\%$ \\
$\modelcomponent_\text{scope}$  & $-2.64$ & $88\%$ \\
\bottomrule
\end{tabular}
\caption{Necessity and sufficiency of the full circuit, $\circuit$, relative to the scope, $\modelcomponent_\text{scope}$, tested on an evaluation set of 192 PP items. The circuit is nearly as necessary and sufficient as the full scope, despite containing around 1/6000 as many MLP neurons.}
\label{tab:necessity-circuit-vs-floor}
\vspace{-5pt}
\end{table}

Importantly, we do not see the circuit as the \emph{complete} agreement mechanism for \model{} --- \Cref{fig:head-scan} and \Cref{fig:circuit-emergence} show that $\layerhead{15}{2}$ and preposition neurons also carry agreement signal --- nor is it constructed to maximize any metric. We interpret the circuit to be a sparse representation of how the model transports the number feature from the subject to the prediction position via a specific $\layerhead{13}{7}$ route. The scope of the circuit $\modelcomponent_\text{scope}$ thus consists of L1--L12 MLPs at $\pos{1}$, \layerhead{13}{7}, and L14--15 MLPs at $\pos{4}$.\footnote{We omit the subject L0 MLP from the scope as MLP ablation there is globally destructive. We take this as evidence that the L0 MLP may be responsible for \emph{constructing} the number feature with our circuit responsible for \emph{transporting} the feature to readout. Full details regarding scope selection and L0 MLP ablation results are in App.~\ref{app:scope}.}

\paragraph{Circuit components.}
The circuit seems to carry out the minimal mechanism laid out in \Cref{sec:intro}.
(1) \emph{Amplifier} neurons $\amps$ at the subject process and propagate the number feature, (2) an attention head $\head$ transports this information to the prediction position, and (3) the head outputs and readout neurons $\pros$ and $\antis$ license the corresponding output form.

Our characterization of the neurons is as follows: neurons in $\amps$ have high activations on plural subjects, while their activations on singular inputs are comparable to those on non-nouns --- we thus think of $\amps$ as \emph{plural detectors}. Readout neurons come in two flavors: \emph{pro-agreement} neurons $\pros$ promote the correct verb continuations, while \emph{anti-agreement} neurons $\antis$ seem to oppose agreement. For the complete set of circuit components and more details on this characterization, see App.~\ref{app:circuit-details}.

\paragraph{Evaluating the circuit.}
% \sasha[]{The notation feels super overloaded in this section, we have multiple different versions of C, many different G's... can it be simplified?}
We now measure the necessity and sufficiency of our final circuit, providing the same for the entire scope as a baseline in \Cref{tab:necessity-circuit-vs-floor}. We find our circuit is $78\%$ necessary, with its mean ablation dropping the agreement gap nearly as much as mean ablating the full scope. We also find our circuit is $83\%$ sufficient, which is comparable to the $88\%$ sufficiency of the entire scope. Further, the circuit's complement has little effect on agreement (App.~\ref{app:off-circuit}).
%We do note, however, that there is some remaining effect even after ablating the full scope we consider. This suggests that the mechanism is more distributed than 
% 
% We further demonstrate that the circuit complement relative to the scope has near-zero necessity and sufficiency, confirming that the effect is specific to the circuit rather than arbitrary activations along the path (with the relevant details of this experiment in \Cref{tab:intervention-off-circuit}).

\begin{takeaway}
A sparse circuit of 18 neurons and a single attention head captures nearly all of the agreement information flow within the scope.
\end{takeaway}

\subsection{Characterizing the final circuit}

Having produced and validated a sparse circuit on \textsc{PP} templates, we now show that it mediates agreement more generally.
Then, we demonstrate that we can compress the sparse circuit even further, localizing the information flow within $\head$ to a one-dimensional subspace. 

\paragraph{Generalization.}\label{sec:generalization}

First, we show our circuit is necessary and sufficient for agreement in further settings. In particular, we measure our circuit on three synthetic templates with novel structure --- agreement across a subject relative-clause (\textsc{SubjectRC}), an object relative-clause (\textsc{ObjectRC}), or no distractors (\textsc{Simple}) --- as well as on mined natural examples from WikiText 
\citep{merity2016pointer} for which we manually construct the minimal pair. 
Templates and exemplars for each evaluation are in App.~\ref{app:templates}.

% First, we show that our circuit is necessary and sufficient for agreement across synthetic templates. These consist of two templates similarly across distractors --- a subject- or object-relative clause (\textsc{SubjectRC}, \textsc{ObjectRC}) --- and a \textsc{Simple} agreement template where the subject and the verb are adjacent.

% We also evaluate the circuit on naturalistic sentences. To this end, we find 80 naturally occurring examples of number agreement across distractors (40 singular subjects, 40 plural subjects) in WikiText 
% \citep{merity2016pointer}. We hand-label these entries by subject number, and create minimal pairs by replacing each subject and predicted verb with their opposite-number form.

\begin{table}[t]
    \centering
    \small
%     \begin{tabular}{@{}l p{0.68\columnwidth}@{}}
%     \toprule
%     Template & Example minimal pair \\
%     \midrule
%     \textsc{PP} & \textit{The guard(s) near the officer} $\rightarrow$ \textit{is/are} \\
%     \midrule
%     \textsc{SubjectRC} & \textit{The guard(s) that liked the officer} $\rightarrow$ \textit{is/are} \\
%     \textsc{ObjectRC} & \textit{The guard(s) that the officer liked} $\rightarrow$ \textit{is/are} \\
%     \textsc{Simple} & \textit{The guard(s)} $\rightarrow$ \textit{is/are} \\
%     \midrule
%     WikiText & \textit{The objective(s) of the regulation} $\rightarrow$ \textit{is/are} \\
%     \bottomrule
%     \end{tabular}
% \caption{Examples of the synthetic templates and WikiText sentences.}
% \label{tab:gen-examples}
% \bigskip
\begin{tabular}{l c c c c}
\toprule
\textbf{Template}      & $\mathbf{\gapbase}$  & $\mathbf{\gapsrc}$ & \textbf{Nec.} & \textbf{Suff.} \\
\midrule
\textsc{PP}                          & $+3.48$ & $-3.48$ & $78\%$ & $83\%$  \\ \midrule \midrule
\textsc{SubjectRC}     & $+4.48$ & $-4.48$ & $86\%$ & $89\%$ \\
\textsc{ObjectRC}      & $+4.30$ & $-4.30$ & $76\%$ & $80\%$ \\
\textsc{Simple}     & $+6.49$  & $-6.49$ & $67\%$ & $78\%$ \\ 
\midrule
WikiText & $+4.29$ & $-3.75$ & $73\%$ & $82\%$ \\
\bottomrule
\end{tabular}
\caption{Circuit necessity and sufficiency across templates ($N = 32$ for \textsc{Simple}, $N = 192$ for other templates, and $N = 80$ for WikiText). Mean activations are computed over the larger training set for \textsc{PP}, and directly on the evaluation set otherwise.}
\label{tab:gen-templates}
\end{table}
% \qy[]{we have the option to drop the examples table and refer to the template section in the appendix}

In \Cref{tab:gen-templates}, we broadly see that our circuit's necessity and sufficiency on these templates are comparable to those for \textsc{PP}. The circuit performs slightly worse on the \textsc{Simple} template, which likely reflects the structural difference of the stimuli. In sum, these results suggest that our circuit is general, and not overfit to just the \textsc{PP} templates from which it was discovered.

We also perform additional generalization tests, showing that circuit activations drawn from one template can control agreement in another, and that the circuit holds under most perturbations to the templates and on novel subjects and verbs.
These results, along with failure cases (the circuit failing on templates with an appended adverb, and \textsc{Simple} being a weak donor for cross-template intervention) are found in App.~\ref{app:generalization-details}.

\paragraph{$\head$ can be localized to a 1D channel.}
% $\head$ is the only channel that connects the upstream and downstream components in $\circuit$. This head alone is $67\%$ necessary and $68\%$ sufficient for agreement.

We can further localize the head transport to a subspace of $\layerhead{13}{7}$'s output-value circuits \citep{elhage2021mathematical} using DAS. To this end, we train three seeds of DAS on the \textsc{PP} training set to successfully localize behavior to a 1D subspace of the head's value output. (See App.~\ref{app:das} for DAS training details.)

\Cref{tab:head-channel} shows the results. We find a 1D subspace which essentially captures the head's total causal effect, and both intervening on and mean ablating the subspace complement barely affect agreement. These results suggest that the head's contribution to the agreement task can be surgically localized to a one-dimensional channel.

\begin{takeaway}
    Our circuit is general despite being discovered on the \textsc{PP} template, mediating agreement robustly across synthetic and naturalistic sentences. Further, the information is routed through a single dimension within the attention head.
\end{takeaway}

\begin{table}[t]
\centering
\small
\begin{tabular}{l c c}
  \toprule
  \textbf{Components} $\mathbf{\modelcomponent}$ & $\mathbf{\gap}$ & \textbf{Measure} \\
  \midrule
  \rowcolor{gray!30} \textit{\textbf{Necessity}} & $\mathbf{\gapmean}$ & \textbf{Nec.} \\
  $\head$ (Full)             & $+1.15$ & $67\%$ \\
  $\head$ (1D subspace)        & $+1.01$ & $70.90 \pm 0.03\%$ \\
  $\head$ (255D Complement)    & $+3.64$ & $-4.50 \pm 0.02\%$ \\
  \midrule
  \rowcolor{gray!30} \textit{\textbf{Sufficiency}} & $\mathbf{\gapintv}$ & \textbf{Suff.} \\
  $\head$ (Full)             & $-1.25$ & $68\%$ \\
  $\head$ (1D subspace)        & $-1.50$ & $71.49 \pm 0.02\%$ \\
  $\head$ (255D Complement)    & $+3.71$ & $-3.28 \pm 0.02\%$ \\
  \bottomrule
\end{tabular}
\caption{The 1D channel identified by DAS has a comparable effect to the entire 256D $\head$. We report the mean necessity and sufficiency $\pm$ standard error across three independent seeds of DAS.}
\label{tab:head-channel}
\end{table}

\section{Conclusion}
\label{sec:discussion}

We believe that the circuit discovered makes good on the mechanistic account of number agreement that we started with.
%, whereby subject number information is identified, tracked, and then decoded at the verb. 
Within the scoped L13 H7-mediated pathway in \model{}, ten subject-position neurons propagate number information, L13 H7 carries it to the prediction position through a one-dimensional causal channel, and eight downstream neurons control the verb readout. 
Agreement circuits in \model{} and the ten \modelsmall{} seeds were diffuse early in training but generally became sparser over time.

We leave it to future work to analyze how and why this transition occurs, although we hypothesize the following: the early circuit emerges stochastically out of a set of features that proved useful for doing the subject--verb agreement task. Then, this diffuse circuit serves as a scaffold for a sparser and more specialized circuit to emerge.
Our hope is that, by granularly studying a single syntactic phenomenon, we can complement wider investigations into LMs' syntactic abilities and thereby better characterize linguistic knowledge in LMs \citep[as called for by][]{futrell2025linguistics}.

\section*{Limitations}
The goal of this paper was to focus on just one linguistic task of subject--verb agreement, but this is also, of course, a limitation. We also focus only on English. Future work could generalize these methods to other constructions, models, and languages. For present purposes, we felt that a wider investigation would have undermined the particular motivation of this project.

Further, we find neuron sparsification and head rerouting to be common across model training runs, but have not provided an explanation for it. Another limitation is that the method for identifying the circuit in Experiment 2 was iterative and involved manual analyses.
We were able to proceed in this way because of the highly focused nature of the work.
But more work would be needed to develop a general procedure for applying it to other tasks and settings.

\section*{Acknowledgments} 
We acknowledge funding from NSF CAREER grant 2339729 to Kyle Mahowald and funding from the Harrington Fellowship to Qing Yao. The work of Tiago Pimentel was funded by the Data Analytics Lab at ETH Z\"urich. We thank Jed Guzelkabaagac, Felix Sarnthein, Rodrigo Casado, Souvik Banerjee, Jan Tempus, and Philip Whittington for comments on this manuscript.

\bibliography{refs,anthology_a,anthology_b}

\appendix

\section{Computation cost}
We use one A40 GPU to run all experiments.
For \model{}, a RelP attribution pass over all MLP neurons at one token position, using 400 examples with batch size 4, takes around 40 seconds of GPU time after model loading. Integrated-gradient variants are slower, about 2 minutes per checkpoint-position.

DAS training uses 400 training examples, batch size 4, 5 epochs, and 1-dimensional interventions. In our runs, this takes around 2 minutes per layer. Forward-pass based validation is comparatively cheap: a single necessity or sufficiency evaluation over an evaluation set takes about 1--2 seconds.

\section{Use of AI assistants}
We used AI assistants to aid in developing code for experiments and analyses. We also used AI for summarizing results and suggesting edits.

\section{Task templates and data}\label{app:templates}
We first introduce our synthetic templates: \textsc{PP}, \textsc{SubjectRC}, \textsc{ObjectRC}, and \textsc{Simple}. Then we discuss extensions of these synthetic templates, both in structure and in lexical items, that are used to further stress-test circuit generalization. Lastly, we show the list of all WikiText \citep{merity2016pointer} agreement sentences. 

All templates share 16 subject pairs (singular/plural forms of the same lemma). The singular forms are the following: \textit{guard, doctor, farmer, author, officer, athlete, manager, senator, consultant, teacher, customer, pilot, executive, minister, actor, architect}. All subjects, objects, and RC verbs are single tokens under the \model{} tokenizer. Base and source sentences always use the singular/plural pair of the same lemma, so they differ only in subject number. We measure model predictions on the verb tokens \{\textit{is}, \textit{was}, \textit{has}\} for singular and \{\textit{are}, \textit{were}, \textit{have}\} for plural; each is a single token.

\subsection{Synthetic templates}\label{app:synthetic-templates-extended}

Below, we detail all synthetic templates used to find and test the circuit. Exact templates and exemplars can be seen in \Cref{tab:template-details}.

\paragraph{\textsc{PP}.} This is the primary template used to find and evaluate our circuit. We draw distractor noun-phrases (NPs) from 16 human-role lemmas, using both singular and plural forms of these distractors. We select from 6 prepositions (\textit{near}, \textit{behind}, \textit{in front of}, \textit{next to}, \textit{across from}, \textit{to the side of}).

\paragraph{\textsc{ObjectRC}.} We use this template to measure generalization. We again draw distractor noun-phrases (NPs) from 16 human-role lemmas, but use only the singular form of these distractors. Within the distractors, we utilize 8 past-tense verbs (\textit{hated}, \textit{injured}, \textit{liked}, etc.).

\paragraph{\textsc{SubjectRC}.} We use this template to measure generalization. Like \textsc{ObjectRC}, distractor noun-phrases (NPs) are drawn from 16 human-role lemmas with only the singular form, and utilize the same set of 8 past-tense verbs.

\paragraph{\textsc{Simple}.} We use this template to measure generalization. We do not sample any distractor NPs for this template as it possesses no distractor, with the position after the subject position being the prediction position.

\paragraph{Adjective insertion.} \textbf{Adjective-on-subject} inserts a single adjective between the determiner and the subject noun, e.g., \textit{The tall guard near the doctors}. The \textbf{adjective-on-object} perturbation instead inserts a single adjective inside the distractor NP, e.g., \textit{The guard near the tall doctors}. There are 16 adjectives total, all applicable to human nouns. They are the following: \textit{tall, old, quiet, busy, new, kind, rich, brave, calm, smart, happy, nice, shy, bold, wise, young}.

\paragraph{Adverb insertion.} The \textbf{adverb} perturbation places a frequency adverb after the distractor object and before the predicted verb, e.g., \textit{The guard near the doctors often}. Unlike the adjective perturbations, this changes the lexical category of the token immediately before the prediction position. It is therefore a stronger test of whether the downstream readout module is tied to the final distractor NP token used in the original PP template. There are 6 adverbs: \textit{always, often, usually, sometimes, rarely, never}.

\paragraph{Prefix insertion.}
The prefix perturbations add material before the subject. The \textbf{single-prefix} condition uses one of the six neutral discourse markers: \textit{Apparently}, \textit{Clearly}, \textit{Surprisingly}, \textit{Obviously}, \textit{In fact}, \textit{For context}; the \textbf{locative-prefix} condition uses one of six neutral locative phrases: \textit{In the park}, \textit{At the office}, \textit{At the hospital}, \textit{At the theater}, \textit{In the garden}, \textit{On the street}. We avoid tensed prefixes such as \textit{Yesterday,} because they can independently affect the distribution over verb forms. These perturbations shift the absolute position of the subject and all downstream tokens while leaving the dependency structure unchanged.

\paragraph{Longer distractor.} This perturbation inserts two hand-designed adjectives in the object NP, e.g., \textit{The guard near the respected senior doctors}. This lengthens the intervening attractor phrase and increases the distance between subject and prediction. There are 6 such adjective pairs: \textit{experienced local},
\textit{respected senior}, \textit{careful young}, \textit{skilled retired}, \textit{friendly new}, \textit{talented junior}.

\begin{table*}
\centering
\small
\begin{tabular}{@{}l p{0.34\linewidth} p{0.26\linewidth}@{}}
\toprule
Condition  & Template prefix & Example prefix \\
\midrule
\textsc{Simple} & \texttt{The \{subject\}} & \textit{The guard} \\
\textsc{SubjectRC}  & \texttt{The \{subject\} that \{verbed\} the \{object\}} & \textit{The guard that liked the doctor} \\
\textsc{ObjectRC} & \texttt{The \{subject\} that the \{object\} \{verbed\}} & \textit{The guard that the doctor liked} \\
\textsc{PP} & \texttt{The \{subject\} \{prep\} the \{object\}} & \textit{The guard near the doctors} \\
\midrule
\textsc{PP} + \textbf{adjective-on-subject}& \texttt{The \{adj\} \{subject\} \{prep\} the \{object\}} & \textit{The tall guard near the doctors} \\
\textsc{PP} + \textbf{adjective-on-object}  & \texttt{The \{subject\} \{prep\} the \{adj\} \{object\}} & \textit{The guard near the tall doctors} \\
\textsc{PP} + \textbf{adverb}  & \texttt{The \{subject\} \{prep\} the \{object\} \{adverb\}} & \textit{The guard near the doctors often} \\
\textsc{PP} + \textbf{single-prefix}  & \texttt{\{prefix\} The \{subject\} \{prep\} the \{object\}} & \textit{Apparently, the guard near the doctors} \\
\textsc{PP} + \textbf{locative-prefix}  & \texttt{\{locative-prefix\} the \{subject\} \{prep\} the \{object\}} & \textit{At the office, the guard near the doctors} \\
\textsc{PP} + \textbf{longer-distractor} & \texttt{The \{subject\} \{prep\} the \{adj1\} \{adj2\} \{object\}} & \textit{The guard near the respected senior doctors} \\
\bottomrule
\end{tabular}
\caption{Agreement templates used in cross-template and perturbation generalization experiments. The object/distractor pool contains both singular and plural forms for \textsc{PP}-based templates, and base and source examples differ only in the subject number. We evaluate $N=192$ pairs ($N=32$ for \textsc{Simple}).}
\label{tab:template-details}
\end{table*}

\subsection{Templates for lexical generalizations}\label{app:generalization-lexical}
The lexical-generalization experiments keep the PP dependency structure fixed and vary the lexical items used either in the subject slot or in the next-token verb label. In all cases, the base/source pair is still a minimal subject-number pair. Concrete examples are in \Cref{tab:lexical-template-details}.

\paragraph{Held-out-subject.} This condition replaces the original subject lexicon with 13 single-token human-role lemmas: \textit{lawyer, engineer, nurse, scientist, driver, artist, musician, dancer, priest, cook, singer, worker, painter}. Both singular and plural forms are single tokens under the \model{} tokenizer. The rest of the template is untouched.

\paragraph{Held-out-verb.} This condition uses the original PP subject, preposition, and distractor pools, but changes the next-token prediction labels. We evaluate 36 single-token third-person singular/plural verb pairs: \textit{runs/run, walks/walk, eats/eat, goes/go, sees/see, knows/know, says/say, gets/get, takes/take, makes/make, comes/come, gives/give, finds/find, tells/tell, thinks/think, feels/feel, writes/write, plays/play, works/work, buys/buy, sells/sell, hears/hear, looks/look, seems/seem, calls/call, means/mean, wants/want, needs/need, loves/love, hates/hate, lives/live, opens/open, closes/close, starts/start, stops/stop, waits/wait}.

\begin{table*}
\centering
\small
\begin{tabular}{@{}l p{0.34\linewidth} p{0.26\linewidth}@{}}
\toprule
Condition & Template / readout & Example \\
\midrule
Held-out subjects & \texttt{The \{held-out-subject\} \{prep\} the \{object\}}; readout is \textit{is/are}, \textit{was/were}, \textit{has/have} & \textit{The lawyer near the doctors} $\to$ \textit{is/are} \\
Held-out verbs & \texttt{The \{subject\} \{prep\} the \{object\}}; readout is the held-out verb-pair set & \textit{The guard near the doctors} $\to$ \textit{runs/run} \\
\bottomrule
\end{tabular}
\caption{Lexical-generalization templates. We test a wider range of subject and verb tokens in the \textsc{PP} template to ensure circuit robustness.}
\label{tab:lexical-template-details}
\end{table*}

\subsection{WikiText agreement cases}
\label{app:wikitext-cases}

We mined 80 agreement sentences from WikiText, balanced across 40 plural-subject and 40 singular-subject cases. For each sentence we manually created a counterfactual that differs in subject number. The subject, counterfactual subject, and verb continuations tokenize as single tokens. When the attested WikiText verb is past tense, we manually convert the verb to its present tense and score the singular/plural pair.

For each item below, \textbf{b} (base) is the attested WikiText prefix plus its attested verb (converted to present-tense, if necessary), and \textbf{s} (source) is the manually constructed opposite-number subject version plus the opposite-number present-tense verb.

\paragraph{Plural-subject cases.}
\begin{enumerate}
\item \textbf{b}: The origins of Imagism are \\ \textbf{s}: The origin of Imagism is
\item \textbf{b}: The characteristics of the sculpture have \\ \textbf{s}: The characteristic of the sculpture has
\item \textbf{b}: The eyes of the monument are \\ \textbf{s}: The eye of the monument is
\item \textbf{b}: The affairs of the CWGC are \\ \textbf{s}: The affair of the CWGC is
\item \textbf{b}: The walls of the nave are \\ \textbf{s}: The wall of the nave is
\item \textbf{b}: The leaders of the Coalition are \\ \textbf{s}: The leader of the Coalition is
\item \textbf{b}: The details of the murder are \\ \textbf{s}: The detail of the murder is
\item \textbf{b}: The reasons behind it are \\ \textbf{s}: The reason behind it is
\item \textbf{b}: The origins of Olaf are \\ \textbf{s}: The origin of Olaf is
\item \textbf{b}: The dimensions of the Tetrarch are \\ \textbf{s}: The dimension of the Tetrarch is
\item \textbf{b}: The objectives of the force are \\ \textbf{s}: The objective of the force is
\item \textbf{b}: The stones of the dolmen are \\ \textbf{s}: The stone of the dolmen is
\item \textbf{b}: The symptoms of AML are \\ \textbf{s}: The symptom of AML is
\item \textbf{b}: The actions of the governor are \\ \textbf{s}: The action of the governor is
\item \textbf{b}: The highlights of the show are \\ \textbf{s}: The highlight of the show is
\item \textbf{b}: The commands of the starter are \\ \textbf{s}: The command of the starter is
\item \textbf{b}: The roles of each deity are \\ \textbf{s}: The role of each deity is
\item \textbf{b}: The effects of antimony are \\ \textbf{s}: The effect of antimony is
\item \textbf{b}: The plots of such works are \\ \textbf{s}: The plot of such works is
\item \textbf{b}: The roles of these area commands are \\ \textbf{s}: The role of these area commands is
\item \textbf{b}: The reasons for going to war are \\ \textbf{s}: The reason for going to war is
\item \textbf{b}: The chances of rapid intensification are \\ \textbf{s}: The chance of rapid intensification is
\item \textbf{b}: The origins of the Knights Hospitaller are \\ \textbf{s}: The origin of the Knights Hospitaller is
\item \textbf{b}: The arrangements for conscientious objectors are \\ \textbf{s}: The arrangement for conscientious objectors is
\item \textbf{b}: The regions of north County Galway and east County Mayo have \\ \textbf{s}: The region of north County Galway and east County Mayo has
\item \textbf{b}: The origins of Hurricane Lorenzo are \\ \textbf{s}: The origin of Hurricane Lorenzo is
\item \textbf{b}: The shows in Japan are \\ \textbf{s}: The show in Japan is
\item \textbf{b}: The heads in Lisbon are \\ \textbf{s}: The head in Lisbon is
\item \textbf{b}: The characteristics of its body, head, fins, and teeth are \\ \textbf{s}: The characteristic of its body, head, fins, and teeth is
\item \textbf{b}: The levels of ionizing radiation on Io, Europa, and Ganymede are \\ \textbf{s}: The level of ionizing radiation on Io, Europa, and Ganymede is
\item \textbf{b}: The events of the battle are \\ \textbf{s}: The event of the battle is
\item \textbf{b}: The districts of Somerset are \\ \textbf{s}: The district of Somerset is
\item \textbf{b}: The corners of the base have \\ \textbf{s}: The corner of the base has
\item \textbf{b}: The identities of the perpetrator or the instigator are \\ \textbf{s}: The identity of the perpetrator or the instigator is
\item \textbf{b}: The statues in the west tower are \\ \textbf{s}: The statue in the west tower is
\item \textbf{b}: The levels in Perfect Dark have \\ \textbf{s}: The level in Perfect Dark has
\item \textbf{b}: The songs on the album are \\ \textbf{s}: The song on the album is
\item \textbf{b}: The objectives of the German, Austrian and Ottoman advance are \\ \textbf{s}: The objective of the German, Austrian and Ottoman advance is
\item \textbf{b}: The tactics employed by the Anzac Mounted Division are \\ \textbf{s}: The tactic employed by the Anzac Mounted Division is
\item \textbf{b}: The lines about the corners of the eyes and mouth are \\ \textbf{s}: The line about the corners of the eyes and mouth is
\end{enumerate}

\paragraph{Singular-subject cases.}
\begin{enumerate}
\item \textbf{b}: The game of checkers is \\ \textbf{s}: The games of checkers are
\item \textbf{b}: The influence of the churches is \\ \textbf{s}: The influences of the churches are
\item \textbf{b}: The strength of the rearguards is \\ \textbf{s}: The strengths of the rearguards are
\item \textbf{b}: The tail behind the stings has \\ \textbf{s}: The tails behind the stings have
\item \textbf{b}: The family of four is \\ \textbf{s}: The families of four are
\item \textbf{b}: The head of a given divine image is \\ \textbf{s}: The heads of a given divine image are
\item \textbf{b}: The number of pilots licensed by the CAA to fly powered aircraft in 2005 is \\ \textbf{s}: The numbers of pilots licensed by the CAA to fly powered aircraft in 2005 are
\item \textbf{b}: The objective of regulation is \\ \textbf{s}: The objectives of regulation are
\item \textbf{b}: The goal of the blockade is \\ \textbf{s}: The goals of the blockade are
\item \textbf{b}: The attack on Ancona is \\ \textbf{s}: The attacks on Ancona are
\item \textbf{b}: The objective of the bombardment of Ancona is \\ \textbf{s}: The objectives of the bombardment of Ancona are
\item \textbf{b}: The application of the Johnson-Corey-Chaykovsky reaction in organic synthesis is \\ \textbf{s}: The applications of the Johnson-Corey-Chaykovsky reaction in organic synthesis are
\item \textbf{b}: The tone of these chapters is \\ \textbf{s}: The tones of these chapters are
\item \textbf{b}: The structure of Dominican society is \\ \textbf{s}: The structures of Dominican society are
\item \textbf{b}: The panel to the east of the north portico is \\ \textbf{s}: The panels to the east of the north portico are
\item \textbf{b}: The door of the central shrine has \\ \textbf{s}: The doors of the central shrine have
\item \textbf{b}: The state of Polish primary schools is \\ \textbf{s}: The states of Polish primary schools are
\item \textbf{b}: The inclusion of occupational debris over the bones is \\ \textbf{s}: The inclusions of occupational debris over the bones are
\item \textbf{b}: The use of Sb as the standard chemical symbol for antimony is \\ \textbf{s}: The uses of Sb as the standard chemical symbol for antimony are
\item \textbf{b}: The origin of this is \\ \textbf{s}: The origins of this are
\item \textbf{b}: The formation of halogenated antimony compounds is \\ \textbf{s}: The formations of halogenated antimony compounds are
\item \textbf{b}: The budget of the film is \\ \textbf{s}: The budgets of the film are
\item \textbf{b}: The cost of construction is \\ \textbf{s}: The costs of construction are
\item \textbf{b}: The peak of the roof is \\ \textbf{s}: The peaks of the roof are
\item \textbf{b}: The method and timing of mowing is \\ \textbf{s}: The methods and timing of mowing are
\item \textbf{b}: The influence of weather on chick survival is \\ \textbf{s}: The influences of weather on chick survival are
\item \textbf{b}: The goal of induction therapy is \\ \textbf{s}: The goals of induction therapy are
\item \textbf{b}: The island of Hispaniola, shared by Haiti and the Dominican Republic, is \\ \textbf{s}: The islands of Hispaniola, shared by Haiti and the Dominican Republic, are
\item \textbf{b}: The consequence of failing a puzzle of this sort is \\ \textbf{s}: The consequences of failing a puzzle of this sort are
\item \textbf{b}: The relationship between the tolling points and interchanges along SR 878 and SR 874 is \\ \textbf{s}: The relationships between the tolling points and interchanges along SR 878 and SR 874 are
\item \textbf{b}: The purpose of this plan is \\ \textbf{s}: The purposes of this plan are
\item \textbf{b}: The amount of water in a toilet or sink is \\ \textbf{s}: The amounts of water in a toilet or sink are
\item \textbf{b}: The form of the poem is \\ \textbf{s}: The forms of the poem are
\item \textbf{b}: The rate of protein synthesis is \\ \textbf{s}: The rates of protein synthesis are
\item \textbf{b}: The study of proteins in vivo is \\ \textbf{s}: The studies of proteins in vivo are
\item \textbf{b}: The field of bioinformatics is \\ \textbf{s}: The fields of bioinformatics are
\item \textbf{b}: The task allocated to the British Commonwealth force is \\ \textbf{s}: The tasks allocated to the British Commonwealth force are
\item \textbf{b}: The definition of leg before wicket is \\ \textbf{s}: The definitions of leg before wicket are
\item \textbf{b}: The object on the right seated on legs alongside a box is \\ \textbf{s}: The objects on the right seated on legs alongside a box are
\item \textbf{b}: The song, ``a cheery little tune about death, destruction and the end of the world'' is \\ \textbf{s}: The songs, ``a cheery little tune about death, destruction and the end of the world'' are
\end{enumerate}

% \subsection{Templates for verb-specificity}\label{app:generalization-non-verbs}
% The verb-specificity control uses the same PP prefixes and subject-number base/source pairs as the held-out-verb experiment, but replaces the agreement-verb readout with unpaired non-verb tokens. This tests whether the circuit intervention broadly perturbs next-token probabilities or specifically shifts agreement-relevant verb probabilities.

% We use 20 high-frequency single-token non-verbs under the \model{} tokenizer: \textit{book, table, time, water, money, friend, story, day, year, place, hand, life, idea, family, country, system, problem, room, child, school}. These tokens are evaluated independently rather than as singular/plural pairs.

% \begin{table*}[h]
% \centering
% \small
% \begin{tabular}{@{}l p{0.34\linewidth} p{0.26\linewidth}@{}}
% \toprule
% Condition & Template / readout & Example \\
% \midrule
% Non-verb controls & \texttt{The \{subject\} \{prep\} the \{object\}}; readout is each non-verb token independently & \textit{The guard near the doctors} $\to$ \textit{book}, \textit{table}, \textit{time} \\
% \bottomrule
% \end{tabular}
% \caption{Verb-specificity control template. The prefix distribution is identical to the PP lexical-generalization setting, but the measured tokens are non-verbs rather than verb forms.}
% \label{tab:verb-specificity-template-details}
% \end{table*}

\section{Additional methodological details}\label{app:methods-details}
% Mirrors Sec.~\ref{sec:model-and-task}. This section should collect supporting
% details for the task/data, neuron attribution, and routing methods, in the
% order in which they are introduced in the main Methods section.

% Referenced from ``Subject--verb agreement task'' in Sec.~\ref{sec:model-and-task}
% via App.~\ref{app:templates}. It should contain the complete synthetic-template
% definitions, lexical inventories, tokenization constraints, and construction
% of the disjoint discovery and evaluation sets. Move the existing ``Template
% Details'' section here together with its label app:templates.

% We supplement the methods section here.

\subsection{RelP neuron attribution}\label{app:attribution-details}
% Referenced from ``Neuron identification'' in Sec.~\ref{sec:methods} via
% App.~\ref{app:attribution-details}. It should contain the replacement-model
% linearization rules, task-specific attribution choices, stop-gradient and
% baseline checks, the integrated-gradients comparison, and position-pooling
% details. Move the existing ``Neuron Identification'' section here together
% with its label app:attribution-details.

We apply RelP \citep{jafari2025relp, arora2026languagemodelcircuitssparse} to form a broad candidate set of MLP neurons before running causal validation. The attribution score is computed over the PP training set ($N = 400$).

\paragraph{Replacement-gradient attribution.}
For each layer $L$ and neuron $n$, we capture the post-nonlinearity MLP activation $v_{L,n}$ at the relevant token position and score
\[
    \mathrm{score}_{L,n}
    =
    \mathbb{E}_{(\mathbf{b},\mathbf{s})}
    \left[
        v_{L,n}(\mathbf{b})
        \frac{\partial m(\mathbf{b}, \mathbf{s})}{\partial v_{L,n}(\mathbf{b})}
    \right],
\] where $m(\mathbf{b}, \mathbf{s}) = \log (p(\ell_\mathbf{b} \mid \mathbf{b})) - \log (p(\ell_\mathbf{s} \mid \mathbf{b}))$ is the per-example agreement gap.

Linearizations are architecture specific. For Pythia, we use the following locally linear replacement model: LayerNorm centering and scaling factors are frozen, GELUs are replaced by their local linear gates, and attention softmax probabilities are detached before the value-weighted sum.

There are two implementation choices which deviate from
\citet{arora2026languagemodelcircuitssparse} --- both of which we show do not qualitatively affect our results. First, the score in \Cref{eq:relp} uses the zero-baseline form $v_n\cdot\nabla$, which \citet{arora2026languagemodelcircuitssparse} adopt in the unpaired setting, rather than the activation-difference form $\Delta v_n\cdot\nabla$ they use with paired data. The former estimates a neuron's contribution on the base input, while the latter estimates the effect of activation replacement from the counterfactual. Zero-baseline is a computationally simpler method since it does not require a source-activation pass and storage. This choice has little impact on the candidates: comparing both the zero-baseline form $v_n\cdot\nabla$ and the activation-difference form $\Delta v_n\cdot\nabla$, we find that they rank the top $50$ neurons identically and overlap on $944$ of the top $1024$ neurons pooled across positions. This suggests our findings are largely unaffected by the choice of zero-baseline.

Second, we also detach the outputs of all other MLPs when scoring a neuron within a particular layer $L$, so the gradient reaching $L$ flows only through the residual stream and attention. \citet{arora2026languagemodelcircuitssparse} introduce this for edge attribution, but we also apply it to node scoring here. There is no clear unbiased choice here: our approach will underweight neurons whose contributions depend on downstream MLP processing, while the alternative can give credit to both upstream and downstream neurons that mediate the same effect.

We compared the two approaches and found that the broader ranking changes, but the conclusion does not. We recomputed the top neuron candidates for \model{} with the stop-gradient removed (i.e., total-effect scoring, where the gradient can flow through downstream MLPs). The two methods overlap in 42 of the top 50 ranking neurons, though this overlap falls to 356 at top-1024. However, the final circuit's 18 neurons are all within the top-35 in both rankings, and the sparsification claim of \Cref{sec:behavior-vs-mechanism} does not change.

\Cref{tab:stopgrad-robustness} shows that the portion of necessity or sufficiency recovered from adding the top-16 neurons to the head-baseline remains low early in training and high later in training, regardless of direct- or total-effect scoring.

\begin{table}
\centering
\small
\begin{tabular}{l l c c c c}
\toprule

 & & \multicolumn{4}{c}{Percent Recovered} \\ \cmidrule(lr){3-6}
 & & \multicolumn{2}{c}{Full pool} & \multicolumn{2}{c}{Top-16} \\
\cmidrule(lr){3-4}\cmidrule(lr){5-6}
Step & Scoring & Nec & Suff & Nec & Suff \\
\midrule
\multirow{2}{*}{2k}   & Direct (used)  & $89.4$ & $92.7$ & $12.2$ & $17.8$ \\
                      & Total effect   & $88.8$ & $92.5$ & $10.8$ & $16.2$ \\
\midrule
\multirow{2}{*}{143k} & Direct (used)  & $97.6$ & $98.7$ & $51.7$ & $64.6$ \\
                      & Total effect   & $94.9$ & $94.9$ & $54.9$ & $66.3$ \\
\bottomrule
\end{tabular}
\caption{Sparsification is robust to the stop gradient implementation. All values are percentages on the \textsc{PP} evaluation set. ``Full pool'' is the necessity/sufficiency of the 1024-neuron circuit with respective attention heads, and ``Top-16'' is the fraction of the gain from the head-only baseline to the full pool captured by the 16 highest-ranked neurons.}
\label{tab:stopgrad-robustness}
\end{table}

\paragraph{Integrated-gradient check.}
As a robustness check during candidate generation, we also compute integrated gradients on MLP activations \citep{sundararajan2017axiomatic, marks2025sparse}. This method interpolates the captured MLP activation from the counterfactual source activation to the base activation, keeps all other activations on the base run, and averages gradients along a $k$-step path. The resulting score is
\[
    (v^{\mathrm{base}}_{L,n} - v^{\mathrm{src}}_{L,n})
    \cdot
    \frac{1}{k}\sum_{i=1}^{k}
    \frac{\partial m}{\partial v^{(i)}_{L,n}} .
\]
The integrated-gradient scores are used to check that high-ranked candidates are not biased by the local linearizations.

Using $k=8$, we find that at the final checkpoint of \model{}, all $18$ neurons in the final circuit fall within the top $35$ neurons under both pooled RelP rankings and integrated gradients, and the two agree on $44$ of the top $50$ neurons.

\subsection{Neuron--head routing attribution}\label{app:routing-attribution-details}
% Referenced from ``Neuron and head routing'' in Sec.~\ref{sec:methods} via
% App.~\ref{app:routing-attribution-details}. It should define the upstream and
% downstream restoration tests, routing scores, visualization thresholds, and
% sensitivity analyses. Move the existing ``Routing Attribution Details''
% section here together with its label app:routing-attribution-details.

At each checkpoint, we consider the top-32 positive RelP neurons at the subject, preposition, and prediction positions. In \Cref{sec:circuit-comparison}, we ask whether these neurons route through the early $\layerhead{15}{7}$ pathway, the later $\layerhead{13}{7}$ pathway, a mixture of the two, or neither.

Our detailed methodology is as follows. For an upstream neuron $n$, we first compute its sufficiency in isolation. Next, we repeat the intervention, but this time hold one downstream head $h$ at its base value (for the edge between the neuron's position and the prediction position). This manipulation blocks $n$'s effect from propagating through $h$. We denote the neuron's \emph{routed sufficiency} via $h$ as:
\[
\operatorname{RSuff}_{n \rightarrow h} = \operatorname{Suff}(n) - \operatorname{Suff}(n; \operatorname{block}(h)).
\]
The routed sufficiency value denotes the portion of the neuron's effect that disappears once $h$ is blocked.

Similarly, for a downstream neuron, $n$, and an upstream head, $h$, we first intervene on the head, then repeat the intervention whilst restoring the neuron $n$'s activation.
\[
\operatorname{RSuff}_{h \rightarrow n} = \operatorname{Suff}(h) - \operatorname{Suff}(h; \operatorname{restore}(n)).
\]
This measures how much the head intervention effect is mediated through $n$.

\paragraph{Visualization.}
We now explain how we colored neurons in \Cref{fig:circuit-emergence}. These computations and thresholds are purely for visualization purposes.
%and are not ``canonical'' in any way. 

We first identify the neurons on either head's route. This takes different forms for upstream and downstream neurons. For an upstream neuron to qualify, we jointly block both heads, and require the neuron to lose at least $5\%$ of the standalone sufficiency. For a downstream neuron, we require restoring the activation to remove at least $5\%$ of the joint sufficiency from an intervention on both heads. If a neuron does not meet these thresholds, it is considered to not be on either head's route. 

For the neurons on either head's routes, we attribute their colors as follows. We first define
\[
  r_{13}(n)=
  \begin{cases}
  \operatorname{RSuff}_{n \rightarrow 13}, & n\text{ is upstream},\\
  \operatorname{RSuff}_{13 \rightarrow n}, & n\text{ is downstream},
  \end{cases}
  \] and define $r_{15}(n)$ likewise. This allows us to define positive routing shares
\begin{align*}
  s_{13}(n) &=
  \frac{[r_{13}(n)]_+}
       {[r_{13}(n)]_+ + [r_{15}(n)]_+}, \\
  s_{15}(n) &= 1-s_{13}(n),
\end{align*} where $[x]_+$ denotes $\max(x, 0)$. 

This metric gives us a gradient between how much a neuron is $\layerhead{15}{7}$-routed or $\layerhead{13}{7}$-routed. Neurons are duly colored by this gradient: blue when $s_{15} = 1$, purple when the two shares are equal, and red when $s_{13} = 1$. %Since $\layerhead{15}{7}$ has no downstream neurons, any prediction-side neuron must be either uncolored or red.

\section{Experiment 1: Additional results}\label{app:exp1-details}
% Mirrors Sec.~\ref{sec:emergence}. This section should collect the analyses
% supporting neuron sparsification, persistent neuron membership, exceptional
% training trajectories, and attention-head behavior across checkpoints/seeds.
\subsection{Different levels of $k$}
\begin{figure*}
    \centering
    \centering
  \begin{minipage}{0.48\textwidth}
    \centering
    \includegraphics[width=\linewidth]{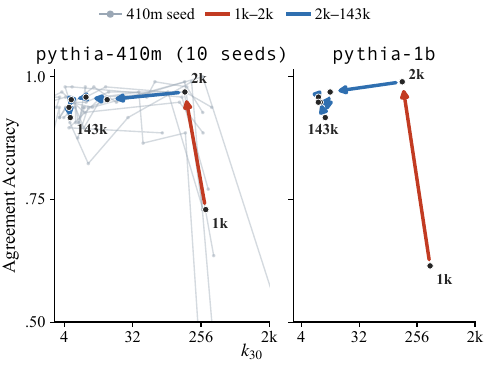}
    \caption{Agreement accuracy vs. $k_{30}$ over training checkpoints.}
    \label{fig:k30}
  \end{minipage}
  \hfill
  \begin{minipage}{0.48\textwidth}
    \centering
    \includegraphics[width=\linewidth]{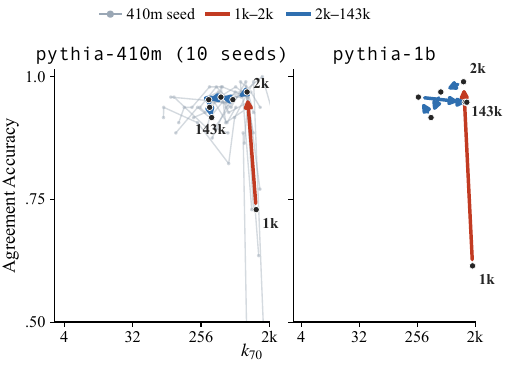}
    \caption{Agreement accuracy vs. $k_{70}$ over training checkpoints.}
    \label{fig:k70}
  \end{minipage}
\end{figure*}

We also produce \Cref{fig:behavior-vs-mechanism} at different threshold levels. Define $k_{30}$ and $k_{70}$ similarly to eq.~\ref{eq:k50}, we reproduce \Cref{fig:behavior-vs-mechanism} in \Cref{fig:k30} and \Cref{fig:k70}, respectively.

We see that broadly speaking, the number of neurons in these sets decrease over training. For $k_{70}$ in \model{}, we see a rebound at step 143k. This is likely due to the final circuit of \model{} relying on downstream prediction-position readout neurons, which we find in Experiment 2.

We take these results to be consistent with the trend that causal effects of subject neurons becomes more concentrated over training. This notion of sparsification may not always be reflected in our figures, and building a complete circuit requires examining model- and checkpoint-specific mechanisms beyond the subject position.

\subsection{Overlap of neuron sets across training}\label{app:neuron-overlap}
For each checkpoint, we report each model's $k_{50}$ set size, and report the overlap between consecutive evaluated checkpoints' $k_{50}$ sets. Since this set does not monotonically decrease, we ``normalize'' by also showing the smaller size of two consecutive sets.

\begin{table*}[t]
      \centering
      \small
      \setlength{\tabcolsep}{3pt}
      \renewcommand{\arraystretch}{1.12}
      \begin{tabular}{lcccccc}
      \toprule
      Run
      & 2k
      & 5k
      & 8k
      & 18k
      & 50k
      & 143k
\\
      \midrule
      Seed 0
      & 766
      & 152 [108/152]
      & 55 [53/55]
      & 15 [15/15]
      & 14 [11/14]
      & 12 [12/12]
 \\
      Seed 1
      & 462
      & 170 [102/170]
      & 85 [64/85]
      & 22 [22/22]
      & 12 [11/12]
      & 6 [6/6]
 \\
      Seed 2
      & 345
      & 153 [81/153]
      & 39 [36/39]
      & 29 [28/29]
      & 23 [21/23]
      & 48 [23/23]
 \\
      Seed 5
      & 477
      & 319 [136/319]
      & 311 [182/311]
      & 178 [126/178]
      & 58 [50/58]
      & 29 [29/29]
 \\
      Seed 6
      & 569
      & 198 [118/198]
      & 109 [95/109]
      & 15 [15/15]
      & 9 [9/9]
      & 8 [8/8]
 \\
      Seed 7
      & 473
      & 90 [66/90]
      & 29 [28/29]
      & 17 [17/17]
      & 16 [14/16]
      & 14 [13/14]
\\
      Seed 8
      & 467
      & 114 [73/114]
      & 42 [40/42]
      & 14 [12/14]
      & 15 [12/14]
      & 14 [14/14]
 \\
      Seed 9
      & 472
      & 154 [91/154]
      & 47 [45/47]
      & 24 [24/24]
      & 36 [22/24]
      & 58 [34/36] 
\\
      \midrule
      Seed 3 (outlier)
      & 491
      & 68 [52/68]
      & 35 [35/35]
      & 14 [14/14]
      & 12 [10/12]
      & 30 [0/12]
\\
      Seed 4 (outlier)
      & 446
      & 357 [119/357]
      & 349 [183/349]
      & 160 [108/160]
      & 35 [35/35]
      & 255 [1/35]
 \\
      \midrule
      \model{}
      & 555
      & 83 [70/83]
      & 42 [40/42]
      & 21 [21/21]
      & 41 [20/21]
      & 31 [29/31]
\\
      \bottomrule
      \end{tabular}
    \caption{Size and overlap of the $k_{50}$ neuron sets across checkpoints. The first number is the $k_{50}$ size, the numerator of the fraction is the overlap between the checkpoint's and the previous checkpoint's $k_{50}$ set, and the denominator of the fraction is the size of the smaller $k_{50}$ set of the two.}
      \label{tab:neuron-persistence}
\end{table*}

Our results are in \Cref{tab:neuron-persistence}. We see that generally, latter checkpoint's neuron sets are largely subsets of the previous checkpoint's neuron sets. In the rare case that $k_{50}$ counts expand (possibly due to instabilities in how they are estimated), the neurons belonging to the smaller set are largely retained. Outliers show a qualitatively different pattern entirely, with abrupt changes in their neuron set memberships.

\subsection{Late-training outlier trajectories}\label{app:outlier}
\begin{figure}
    \centering
    \includegraphics[width=0.98\linewidth]{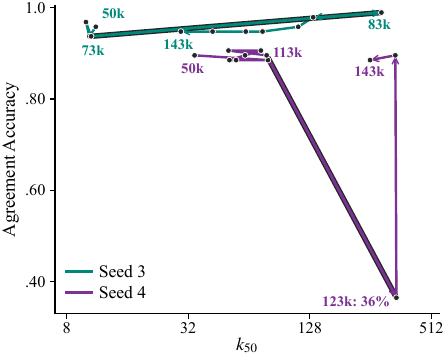}
    \caption{$k_{50}$ rebound co-occurs with training regressions. Seed 3 rebound happens early and the model recovers a sparse neuron mechanism by end of training. Seed 4 suffered setbacks late and its $k_{50}$ ended high.}
    \label{fig:outlier}
\end{figure}

The pattern of models first learning behavior before the internal mechanism sparsifies seems to hold across non-outlier seeds. However, the outlier seeds are interestingly different --- \Cref{fig:outlier} shows that at exactly the training setbacks reported by \citet{vanderwal2025PolyPythias}, the outlier models' $k_{50}$ scores rebound. For seed 3, the setback is not on agreement behavior but only on the mechanism, so subsequent training simply re-sparsifies its neuron mechanism. However, seed 4 loses a significant portion of its agreement abilities. As such, in the remaining training steps, it relearns agreement, but runs out of time for re-sparsification.

Taken together, these results suggest that there is a general pressure to sparsify after agreement is learned, even for outlier runs.

\subsection{Attention-head localization across checkpoints and seeds}\label{app:head-across-seeds}
To identify candidate transport heads, we run a scan on all heads at the prediction position at each checkpoint, testing for both necessity and sufficiency. We rank heads by $\min(\operatorname{Nec}, \operatorname{Suff})$.

For \model{}, we show the top-5 heads at each checkpoint in \Cref{tab:head-scan}, bolding the three heads in the emergence analysis (\layerhead{13}{7}, \layerhead{15}{7}, and \layerhead{15}{2}). We omit early checkpoints because normalizing against a near-zero agreement gap for necessity/sufficiency is unstable. Across the ten seeds, the dominant channel at the end of training is between layers 16--23 (the last layer has index 23). On average, the head settles at a depth of 83\% relative to the total number of layers, comparable to $13/16 \approx 0.81$ in \model{}. This suggests that the late-layer placement of the transport head may not be incidental.

\begin{table}
\centering
  \small
  \begin{tabular}{l l r}
  \toprule
  Checkpoint & Head & $\min\{\mathrm{Nec},\mathrm{Suff}\}$ \\
  \midrule
  1k
   & \textbf{\layerhead{15}{2}} & $48.4\%$ \\
   & \textbf{\layerhead{15}{7}} & $42.5\%$ \\
   & \layerhead{15}{0} & $8.2\%$ \\
   & \layerhead{15}{1} & $7.8\%$ \\
   & \layerhead{9}{6}  & $1.7\%$ \\
  \addlinespace
  2k
   & \textbf{\layerhead{15}{7}} & $49.4\%$ \\
   & \textbf{\layerhead{15}{2}} & $30.7\%$ \\
   & \layerhead{15}{0} & $7.2\%$ \\
   & \layerhead{9}{6}  & $3.6\%$ \\
   & \layerhead{7}{6}  & $3.5\%$ \\
  \addlinespace
  5k
   & \textbf{\layerhead{15}{7}} & $51.2\%$ \\
   & \textbf{\layerhead{15}{2}} & $24.2\%$ \\
   & \textbf{\layerhead{13}{7}} & $11.5\%$ \\
   & \layerhead{15}{0} & $2.3\%$ \\
   & \layerhead{7}{6}  & $1.7\%$ \\
  \addlinespace
  8k
   & \textbf{\layerhead{15}{7}} & $42.8\%$ \\
   & \textbf{\layerhead{13}{7}} & $21.6\%$ \\
   & \textbf{\layerhead{15}{2}} & $18.0\%$ \\
   & \layerhead{7}{6}  & $2.6\%$ \\
   & \layerhead{15}{0} & $1.3\%$ \\
  \addlinespace
  18k
   & \textbf{\layerhead{13}{7}} & $41.1\%$ \\
   & \textbf{\layerhead{15}{7}} & $24.5\%$ \\
   & \textbf{\layerhead{15}{2}} & $15.1\%$ \\
   & \layerhead{7}{6}  & $5.3\%$ \\
   & \layerhead{13}{0} & $2.7\%$ \\
  \addlinespace
  50k
   & \textbf{\layerhead{13}{7}} & $54.9\%$ \\
   & \textbf{\layerhead{15}{2}} & $13.9\%$ \\
   & \textbf{\layerhead{15}{7}} & $8.5\%$ \\
   & \layerhead{7}{6}  & $8.0\%$ \\
   & \layerhead{13}{0} & $3.5\%$ \\
  \addlinespace
  143k
   & \textbf{\layerhead{13}{7}} & $67.1\%$ \\
   & \textbf{\layerhead{15}{2}} & $14.2\%$ \\
   & \layerhead{7}{6}  & $6.4\%$ \\
   & \textbf{\layerhead{15}{7}} & $3.9\%$ \\
   & \layerhead{13}{4} & $2.7\%$ \\
  \bottomrule
  \end{tabular}
\caption{Top five attention heads at each checkpoint scored by $\min(\mathrm{Nec}, \mathrm{Suff})$. The three main heads are bolded.}
\label{tab:head-scan}
\end{table}

\begin{table*}[t]
  \centering
  \small
  \begin{tabular}{llrllr}
  \toprule
  & \multicolumn{2}{c}{Step 2k}
  & \multicolumn{2}{c}{Step 143k} & \\
  \cmidrule(lr){2-3}\cmidrule(lr){4-5}
  Run & Leader & Score & Leader & Score & Movement \\
  \midrule
  Seed 0 & \layerhead{22}{11} & $69.9\%$
          & \layerhead{22}{11} & $65.2\%$ & Stable \\
  Seed 1  & \layerhead{19}{15} & $89.9\%$
          & \layerhead{19}{15} & $80.7\%$ & Stable \\
  Seed 2  & \layerhead{23}{10} & $49.4\%$
          & \layerhead{21}{9}  & $33.9\%$ & Earlier \\
  Seed 5  & \layerhead{20}{15} & $80.4\%$
          & \layerhead{20}{15} & $87.6\%$ & Stable \\
  Seed 6  & \layerhead{21}{6}  & $42.1\%$
          & \layerhead{22}{4}  & $56.2\%$ & Later \\
  Seed 7  & \layerhead{17}{3}  & $68.6\%$
          & \layerhead{17}{3}  & $50.9\%$ & Stable \\
  Seed 8  & \layerhead{22}{7}  & $85.0\%$
          & \layerhead{16}{13} & $48.6\%$ & Earlier \\
  Seed 9  & \layerhead{23}{1}  & $58.6\%$
          & \layerhead{16}{14} & $32.4\%$ & Earlier \\
  Seed 3 (outlier) & \layerhead{23}{11} & $55.6\%$
          & \layerhead{23}{0} & $52.5\%$ & Same-layer \\
  Seed 4 (outlier) & \layerhead{17}{8} & $93.7\%$
          & \layerhead{22}{6} & $71.3\%$ & Later \\
  \bottomrule
  \end{tabular}
  \caption{Leading head across ten seeds of \modelsmall{} at step 2k and step 143k. Score denotes $\min(\operatorname{Nec}, \operatorname{Suff})$. There is no predictable pattern in how the models reorganize the transport channels.}
  \label{tab:tenseed-heads}
  \end{table*}

We similarly track the top heads across seeds of \texttt{pythia-410m} by $\min(\operatorname{Nec}, \operatorname{Suff})$, with the tracked heads and results in \Cref{tab:tenseed-heads}. The composition of the top-three head set changes in every run, suggesting that the primary transport channel is generally unstable across training. The single most dominant head changes in both $\model$ and six seeds of $\modelsmall$. Among such runs, in three seeds the transport head moves from a later layer toward a middle layer, echoing the \model{} pattern. Two seeds, 4 and 6, shift toward a later layer.

We leave for future studies to understand the learning pressures that drive these dynamics.

\section{Experiment 2: Additional results}\label{app:exp2-details}
% Mirrors Sec.~\ref{sec:circuit}. This section should follow the main-text order:
% construction of the final circuit, definition and interpretation of its
% components, causal validation, and generalization.
\subsection{Initial circuit discovery details}\label{app:discovery}
The neuron candidates for our final circuit were those which write toward the next-layer number representation and those with high RelP scores. Here we provide some details on how we identified number representations in the model, and how we pruned to obtain the final circuit.

\paragraph{Distributed alignment search for number.}
We use Distributed Alignment Search \citep[DAS;][]{geiger2024DAS} to test whether the agreement information carried by $\layerhead{13}{7}$ is concentrated in a low-dimensional subspace. To this end, we train DAS, with the default orthogonal alignment map set at one dimension, on the \textsc{PP} training set with learning rate 5e-3 and batch size 4, for 5 epochs. We train at every layer in the subject and prediction positions with five initialization seeds. To score a neuron's output alignment with one representative number-feature subspace, we use the average of the five subspaces after normalization.

\paragraph{Pruning.}
Neuron candidates surfaced by output-alignment or RelP are not necessarily causally important, and it is possible that two candidates can exhibit interaction effects that lower their joint effects. For these reasons, after constructing an initial circuit with promising neurons, we apply leave-one-out validation to prune the circuit.

On the 21-neuron circuit, we remove each remaining neuron and re-evaluate the circuit on the \textsc{PP} training set. We remove the neuron whose deletion had the least effect on circuit necessity and sufficiency. This applied to both neurons with positive and negative agreement effects, as we want to keep strong anti-agreement neurons to better represent the model's mechanism. We repeated this process three times to remove three neurons.
%The exact threshold and stopping point is up to judgment calls.

\subsection{Circuit scope}\label{app:scope}
% Supports ``Pythia-1B's sparse reorganization'' in Sec.~\ref{sec:circuit}.
% It should justify restricting the schematic to the subject, preposition, and
% prediction positions using source-span head patching and position-wise RelP.
% The main text does not currently point here directly; consider adding a
% reference to App.~\ref{app:position-elimination}. Move the existing
% ``Narrowing Down Neuron Positions'' section here with that label.
Our final circuit is presented as a sparse representation of how \model{} carries the number feature from the subject position (from L1 onwards) to the prediction position via $\layerhead{13}{7}$. To this end, we ignore $\pos{0}$ and $\pos{3}$, exclude the L0 MLP from the scope, and ignore the effects of self-attention. We provide justification for these decisions below.
% We omit the subject L0 MLP primarily for the following reason.

% We first show that we can safely ignore neurons at $\pos{0}$ and $\pos{3}$ (which are both the token ``the'') by confirming that no relevant heads transfer information from those positions, and that their RelP attributions are low. We then further demonstrate that the 

\paragraph{Position selection.}

We first show that we can safely ignore neurons at $\pos{0}$ and $\pos{3}$ (which are both the token ``the'') by confirming that no relevant heads transfer information from those positions, and that their RelP attributions are low.

To this end, we first compute each head's sufficiency on the \textsc{PP} training set at each checkpoint. Then, we compute the sufficiency from intervening on the head, but restricted only to the attention between a source token position and the prediction position. The fraction of the edge-restricted sufficiency relative to the full-head sufficiency serves as a measure for the degree of importance for this position.

We show this metric across heads and checkpoints in \Cref{tab:head-source-span-localization}. Position-specific edge patching shows that the early L15 transport heads are almost entirely subject-to-prediction channels. By contrast, $\layerhead{13}{7}$ starts as a weak subject edge but becomes the dominant late channel, reading both from the subject edge and from the preposition edge at 143k. The determiner positions are not accessed by any of these heads.

\Cref{tab:relp-position-checkpoint-max} further shows the maximum absolute RelP score over all MLP neurons in all layers, demonstrating that the estimated effect of neurons at determiner positions $\pos{0}$ and $\pos{3}$ are negligible. This result provides further confirmation that excluding these positions has negligible impact on the final circuit.
  
\begin{table*}
\centering
\small
\begin{tabular}{l l c c c c}
\toprule
Checkpoint & Head & subject $\to$ pred. & prep. $\to$ pred. & det. $\to$ pred. & object $\to$ pred. \\
\midrule
256 & \layerhead{15}{7} & $0.006$ & $0.001$ & $0.000$ & $0.001$ \\
256 & \layerhead{15}{2} & $-0.012$ & $-0.002$ & $0.000$ & $-0.002$ \\
256 & \layerhead{13}{7} & $0.009$ & $0.002$ & $0.000$ & $0.000$ \\
\midrule
1k & \layerhead{15}{7} & $0.391$ & $-0.004$ & $0.000$ & $-0.004$ \\
1k & \layerhead{15}{2} & $0.446$ & $0.000$ & $0.000$ & $0.000$ \\
1k & \layerhead{13}{7} & $0.006$ & $0.000$ & $0.000$ & $0.000$ \\
\midrule
2k & \layerhead{15}{7} & $0.497$ & $0.000$ & $0.000$ & $-0.001$ \\
2k & \layerhead{15}{2} & $0.312$ & $0.000$ & $0.000$ & $0.000$ \\
2k & \layerhead{13}{7} & $0.017$ & $0.001$ & $0.000$ & $0.000$ \\
\midrule
5k & \layerhead{15}{7} & $0.505$ & $0.002$ & $0.000$ & $0.000$ \\
5k & \layerhead{15}{2} & $0.235$ & $0.000$ & $0.000$ & $0.000$ \\
5k & \layerhead{13}{7} & $0.101$ & $0.005$ & $0.000$ & $0.000$ \\
\midrule
8k & \layerhead{15}{7} & $0.426$ & $0.001$ & $0.000$ & $0.002$ \\
8k & \layerhead{15}{2} & $0.175$ & $0.000$ & $0.000$ & $0.001$ \\
8k & \layerhead{13}{7} & $0.186$ & $0.009$ & $0.000$ & $0.002$ \\
\midrule
18k & \layerhead{15}{7} & $0.242$ & $0.001$ & $0.000$ & $0.002$ \\
18k & \layerhead{15}{2} & $0.141$ & $0.002$ & $0.000$ & $0.001$ \\
18k & \layerhead{13}{7} & $0.324$ & $0.041$ & $0.000$ & $0.005$ \\
\midrule
50k & \layerhead{15}{7} & $0.080$ & $0.001$ & $0.000$ & $0.001$ \\
50k & \layerhead{15}{2} & $0.120$ & $0.002$ & $0.000$ & $0.005$ \\
50k & \layerhead{13}{7} & $0.360$ & $0.087$ & $0.001$ & $0.020$ \\
\midrule
70k & \layerhead{15}{7} & $0.059$ & $0.001$ & $0.000$ & $0.001$ \\
70k & \layerhead{15}{2} & $0.134$ & $0.004$ & $0.000$ & $0.009$ \\
70k & \layerhead{13}{7} & $0.363$ & $0.130$ & $0.003$ & $0.016$ \\
\midrule
143k & \layerhead{15}{7} & $0.039$ & $0.001$ & $0.000$ & $0.002$ \\
143k & \layerhead{15}{2} & $0.110$ & $0.005$ & $0.000$ & $0.019$ \\
143k & \layerhead{13}{7} & $0.380$ & $0.184$ & $0.002$ & $0.016$ \\
\bottomrule
\end{tabular}
\caption{Position-specific edge interventions into the prediction position. Entries are fractions of the full head sufficiency recovered by the single-edge intervention. \layerhead{15}{7} and \layerhead{15}{2} are concentrated on subject-to-prediction edges, while \layerhead{13}{7} is a strong subject edge that increasingly uses the preposition-to-prediction edge late in training.}
\label{tab:head-source-span-localization}
\end{table*}

\begin{table*}
\centering
\small
\begin{tabular}{l c c c c c}
\toprule
Checkpoint & $\pos{0}$ initial \textit{The} & $\pos{1}$ subject & $\pos{2}$ prep. & $\pos{3}$ distractor \textit{the} & $\pos{4}$ object/pred. \\
\midrule
256   & $0.00004$ & $0.00025$ & $0.00004$ & $0.00004$ & $0.00031$ \\
1k    & $0.00069$ & $0.00598$ & $0.00038$ & $0.00014$ & $0.00135$ \\
2k    & $0.00064$ & $0.05164$ & $0.00150$ & $0.00051$ & $0.01050$ \\
5k    & $0.00093$ & $0.22402$ & $0.00680$ & $0.00200$ & $0.07530$ \\
8k    & $0.00114$ & $0.28259$ & $0.01666$ & $0.00528$ & $0.15537$ \\
18k   & $0.00392$ & $0.27067$ & $0.06123$ & $0.01214$ & $0.25152$ \\
50k   & $0.00274$ & $0.33274$ & $0.11713$ & $0.01787$ & $0.47838$ \\
70k   & $0.00252$ & $0.31816$ & $0.13967$ & $0.02424$ & $0.47499$ \\
143k  & $0.00353$ & $0.39645$ & $0.23543$ & $0.03269$ & $0.68888$ \\
\bottomrule
\end{tabular}
\caption{Maximum absolute RelP score by \textsc{PP} template token position across checkpoints. Largest RelP scores occur at the subject and prediction-side positions after agreement emerges, while the sentence-initial determiner and the distractor determiner remain small throughout.}
\label{tab:relp-position-checkpoint-max}
\end{table*}

\paragraph{The L0 MLP.}
% Referenced from the circuit-scope footnote in Sec.~\ref{sec:circuit} via
% App.~\ref{app:L0}. It should explain how L0 constructs number information, why
% its DAS direction is not uniquely identified, and why destructive full-MLP
% ablation prevents including L0 in the evaluated circuit scope.

We find ablating the complement of the circuit is unusually destructive at L0. That is, we compare full MLP mean ablation at the subject position for L0 and L5, using the \textsc{PP} subject-position mean as the replacement and measure cross-entropy of the agreement target at the prediction position. Our results can be seen in \Cref{tab:l0-ablation-ce}. On both PP templates and hand-curated WikiText agreement prefixes, L0 mean ablation raises CE by about $0.82$ nats, while L5 mean ablation leaves CE essentially unchanged. For this reason, we only present our circuit relative to a subject-to-prediction path that excludes subject L0. Further, these results suggest that L0 may be responsible for the \emph{construction} of the number feature, with our circuit merely responsible for its \emph{transport} to prediction.

\paragraph{Subject-side self-attention makes little contribution to agreement.}
We find that mean ablating of each layer and head demonstrates that the self-attention contribution at the subject position, $\alpha_h[\pos{1},\pos{1}] W_O^h V_h(x_{\pos{1}})$ is negligible. In particular, we replace such values with the corresponding dataset mean while leaving non-self attention edges untouched. Across the \textsc{PP} dataset this procedure changes the agreement gap from $+3.48$ to $+3.43$, a drop of only $1.5\%$. This supports ignoring attention heads at the subject position. At the prediction position, no single self-attention head under ablation causes a drop of more than $2.5\%$ of the baseline (\layerhead{9}{7}), and heads L14 or later have near-zero mean ablation effects.

\begin{table}
\centering
\small
\begin{tabular}{l l c c c}
\toprule
Dataset & Ablated & clean CE & ablated CE & $\Delta$CE \\
\midrule
\textsc{PP} & L0 & $4.05$ & $4.86$ & $+0.82$ \\
\textsc{PP} & L5 & $4.05$ & $4.03$ & $-0.01$ \\
WikiText  & L0 & $2.08$ & $2.90$ & $+0.83$ \\
WikiText & L5 & $2.08$ & $2.07$ & $-0.01$ \\
\bottomrule
\end{tabular}
\caption{Cross-entropy impact of full-MLP mean ablation at the subject position. CE is measured on the correct agreement target at the prediction position. \textsc{PP} evaluation set $(N=192)$, WikiText $(N=80)$.}
\label{tab:l0-ablation-ce}
\end{table}

\subsection{Circuit components and functional roles}\label{app:circuit-details}
% Referenced from ``The number-agreement circuit'' and ``Understanding the
% circuit'' in Sec.~\ref{sec:circuit} via Table~\ref{tab:circuit} and
% App.~\ref{app:circuit-details}. It should list all circuit members and provide
% the activation, write-direction, vocabulary, and causal evidence supporting
% the amplifier, transport, pro-readout, and anti-readout labels. Move the
% existing ``Circuit Details'' section here with label app:circuit-details.

\begin{table}
\centering
\small
\begin{tabular}{@{}p{0.30\linewidth} p{0.62\linewidth}@{}}
\toprule
Module & Components \\
\midrule
$\amps$ ($\pos{1}$)
& \layerneuron{1}{7988}, \layerneuron{5}{3793}, \layerneuron{8}{6308},
  \layerneuron{9}{5060}, \layerneuron{9}{3101},
  \layerneuron{10}{747}, \layerneuron{10}{6267},
  \layerneuron{11}{6624}, \layerneuron{11}{239},
  \layerneuron{12}{1535} \\
\midrule
$\head$
& \layerhead{13}{7} \\
\midrule
$\pros$ ($\pos{4}$)
& \layerneuron{14}{1160} (pl);
  \layerneuron{14}{4146}, \layerneuron{14}{6888} (sg) \\
\midrule
$\antis$ ($\pos{4}$)
& \layerneuron{14}{6221}, \layerneuron{15}{2511},
  \layerneuron{15}{3382}, \layerneuron{15}{6710} (pl);
  \layerneuron{15}{2305} (sg) \\
\bottomrule
\end{tabular}
\caption{Components of the subject--verb agreement circuit $\circuit$ in \model{}. $\head$ transports the number feature, and $\pros$ and $\antis$ neurons promote or suppress the labeled verb forms.}
\label{tab:circuit}
\end{table}

A full list of circuit components is in \Cref{tab:circuit}, consisting of four modules: $\amps$, $\head$, $\pros$, and $\antis$. Here we describe our interpretation of each module's contribution to the final circuit, as well as the relevant metrics supporting such interpretations.

\paragraph{Amplifiers.}
The upstream neurons in $\amps$ behave primarily as plural detectors. To demonstrate this, we first hand-wrote 95 singular/plural noun pairs, and then used AI assistance to extract words ending in \textit{-s} in the first 5000 rows of WikiText, giving us a total of $447$ singular/plural noun pairs. We also manually generate 144 non-nouns which are single-token verbs, adjectives, adverbs, and function words. We measure all MLP neurons' activations at the subject position for these words. To mirror the template, activations are extracted in the context \texttt{<|endoftext|>The \{word\}}.

The $\amps$ neurons activate similarly, near 0 on singular nouns and non-nouns, but activate strongly on plural nouns. Most of the selected $\amps$ circuit neurons are among the top one or two plural-selective neurons (i.e., largest mean-activation difference between plural and singular inputs) in their layer: \layerneuron{1}{7988}, \layerneuron{5}{3793}, \layerneuron{8}{6308}, \layerneuron{9}{5060}, \layerneuron{11}{6624}, and \layerneuron{12}{1535} are ranked first by plural-selectivity in their respective layers; \layerneuron{9}{3101}, \layerneuron{10}{6267}, \layerneuron{10}{747}, and \layerneuron{11}{239} are also highly ranked. See \Cref{tab:amp-activation-profiles}.

\begin{table}
\centering
\small
\begin{tabular}{l c c c }
\toprule
Neuron & sg mean & pl mean & non-noun mean  \\
\midrule
\layerneuron{1}{7988}  & $-0.117$ & $1.796$ & $-0.102$  \\
\layerneuron{5}{3793}  & $-0.102$ & $3.307$ & $-0.105$  \\
\layerneuron{8}{6308}  & $-0.091$ & $2.831$ & $-0.056$  \\
\layerneuron{9}{5060}  & $-0.108$ & $3.552$ & $-0.090$\\
\layerneuron{9}{3101}  & $-0.085$ & $1.190$ & $-0.068$  \\
\layerneuron{10}{747}  & $-0.106$ & $1.466$ & $-0.049$ \\
\layerneuron{10}{6267} & $-0.117$ & $2.562$ & $-0.018$ \\
\layerneuron{11}{6624} & $-0.121$ & $1.637$ & $-0.016$ \\
\layerneuron{11}{239}  & $-0.103$ & $1.008$ & $-0.077$  \\
\layerneuron{12}{1535} & $-0.082$ & $2.873$ & $0.006$ \\
\bottomrule
\end{tabular}
\caption{Activations for $\amps$ at the subject position. Means are post-activation MLP neuron values. The selected neurons distinguish plural nouns from singular nouns, while singular nouns remain close to non-nouns.}
\label{tab:amp-activation-profiles}
\end{table}

\paragraph{Transport.}
Results from Experiment 1 at the final checkpoint tell us that $\layerhead{13}{7}$ is the most important head for agreement. Further, App.~\ref{app:das} shows that we can localize the number channel to a 1D subspace within the head. Taken together, these results characterize $\head$ as the transport channel for the circuit, moving relevant information from the subject position to the prediction position.

\paragraph{Readout: pro- and anti-agreement neurons.}
The prediction-side neurons are readout components. They are gated by subject number and write directly toward or away from the correct verb-number direction. The two strongest pro-readouts, \layerneuron{14}{1160} and \layerneuron{14}{4146}, have opposite output alignments: \layerneuron{14}{1160} promotes plural verbs such as \textit{are}, \textit{were}, and \textit{have}, while suppressing singulars such as \textit{is}, \textit{was}, and \textit{has}; \layerneuron{14}{4146} has the opposite alignment. 

Their activations also appear to be gated by the upstream subject number: \layerneuron{14}{1160} is active on plural-subject examples and nearly inactive on singular-subject examples, while \layerneuron{14}{4146} is active on singular-subject examples and nearly inactive on plural-subject examples. \layerneuron{14}{6888} is an additional singular-side readout with the same gating pattern. See \Cref{tab:readout-activation-profiles}.

\begin{table}
\centering
\small
\begin{tabular}{l c c c}
\toprule
Neuron & module & sg mean & pl mean \\
\midrule
\layerneuron{14}{1160} & $\pros$ plural & $-0.06$ & $2.88$ \\
\layerneuron{14}{4146} & $\pros$ singular & $1.67$ & $-0.05$ \\
\layerneuron{14}{6888} & $\pros$ singular & $0.95$ & $-0.11$ \\
\midrule
\layerneuron{14}{6221} & $\antis$ plural-gated & $-0.05$ & $0.69$ \\
\layerneuron{15}{2511} & $\antis$ plural-gated & $-0.05$ & $1.36$ \\
\layerneuron{15}{3382} & $\antis$ plural-gated & $-0.08$ & $1.22$ \\
\layerneuron{15}{6710} & $\antis$ plural-gated & $-0.07$ & $0.77$ \\
\layerneuron{15}{2305} & $\antis$ singular-gated & $1.46$ & $-0.10$ \\
\bottomrule
\end{tabular}
\caption{Readout neuron activations at the prediction position. The pro- and anti-readouts are gated by the number state reaching the prediction position: plural-side neurons activate on plural-subject examples, while singular-side neurons activate on singular-subject examples. Each mean is computed over $96$ singular-subject and $96$ plural-subject base sentences in the \textsc{PP} evaluation set ($N = 192$).}
\label{tab:readout-activation-profiles}
\end{table}

The anti-readouts are active components of the same computation but write in the opposite direction from the pro-readouts. For a baseline agreement gap of $3.48$, individually mean ablating anti-agreement neurons increases the gap by $2.7\%$ -- $7.0\%$. Mean ablating all five increases the gap $24.7\%$ relative to baseline ($N=192$).

\subsection{Off-circuit interventions}\label{app:off-circuit}
% Referenced from ``Evaluating the circuit'' in Sec.~\ref{sec:circuit} via
% Table~\ref{tab:intervention-off-circuit}. It should contain the complement and
% donor-control interventions used to establish that the selected circuit,
% rather than arbitrary activations in its scope, carries the measured effect.
% Move the relevant material from ``Necessity: Additional Details'' here,
% retaining app:necessity if that broader label is still needed.

As an additional robustness test, we intervene on non-circuit MLP activations within the declared scope. Here, we intervene on all MLP blocks within the scope (L1--L12 MLPs at $\pos{1}$ and L14--L15 MLPs at $\pos{4}$) except the circuit neurons.

For each base sentence $\mathbf{b}$ in the \textsc{PP} evaluation set, we intervene on $\mathbf{b}$ with $20$ randomly sampled donor PP sentences and measure the agreement gap. If the complement does not participate in agreement, the post-intervention agreement gap should also show little to no sensitivity to the donor's number. Indeed, \Cref{tab:intervention-off-circuit} shows that the circuit-complement has little effect on agreement under intervention.

\begin{table}
\centering
\small
\begin{tabular}{l c c c}
\toprule
Donor               & all & sg & pl \\
\midrule
Baseline & $+3.48$ & $+3.61$ & $+3.35$  \\
Same-number donor   & $+3.68$ & $+3.71$ & $+3.66$  \\
Cross-number donor  & $+3.54$ & $+3.59$ & $+3.48$  \\
\bottomrule
\end{tabular}
\caption{Agreement gap under off-circuit intervention ($N = 192$, $N_\text{donors}=20$ each; 1924 same-class and 1916 cross-class trials). With the circuit untouched, replacing relevant non-circuit activation with a donor's activation leaves agreement intact regardless of donor class.}
\label{tab:intervention-off-circuit}
\end{table}

\subsection{Generalization datasets and additional results.}\label{app:generalization-details}
% Referenced from ``Generalization'' in Sec.~\ref{sec:circuit}; the final
% paragraph currently points to App.~\ref{app:templates}, although the described
% cross-template, perturbation, lexical, failure-case, and WikiText results are
% in App.~\ref{app:generalization-details}. This subsection should collect those
% experiments and the naturalistic examples. Consider retargeting the main-text
% reference to app:generalization-details, while leaving core template and data
% definitions under the Methods appendix.
In the main text, we have shown that the circuit generalizes to other synthetic templates and naturalistic sentences. We further show that the final circuit is a robust agreement mechanism by showing that the circuit (1) can be used to control agreement across templates, (2) applies to most perturbations to the \textsc{PP} template, with the exception of adding an adverb before predicting the next verb, and (3) applies to subjects and verbs in general.

\paragraph{Circuit activations are transferable.}
The circuit controls agreement across templates, so we expect that activations on the modules are transferable across templates. We test this by intervening on the circuit $\circuit$ across templates, and measuring sufficiency.

For each pair of templates, we create $N = 192$ pairs ($N = 32$ for \textsc{Simple}) differing in subject number drawn from their evaluation sets. We intervene on circuit components, and measure sufficiency. \Cref{tab:cross-template-matrix} contains our results.

\begin{table}
\centering
\small
\begin{tabular}{l c c c c}
\toprule
        & \multicolumn{4}{c}{Donor} \\
\cmidrule(lr){2-5}
Base    & \textsc{PP}      & \textsc{SubjRC}  & \textsc{ObjRC}   & \textsc{Simple}  \\
\midrule
\textsc{PP}      & $\mathbf{84\%}$  & $97\%$  & $95\%$  & $59\%$ \\
\textsc{SubjRC}  & $78\%$  & $\mathbf{89\%}$  & $88\%$  & $58\%$ \\
\textsc{ObjRC}   & $72\%$  & $82\%$  & $\mathbf{80\%}$  & $53\%$ \\
\textsc{Simple}  & $85\%$  & $93\%$  & $90\%$  & $\mathbf{78\%}$ \\
\bottomrule
\end{tabular}
\caption{Sufficiency from cross-template intervention. Rows are the base template; columns are the donor template (each pair drawn at random, with the donor's subject opposite-number from base's). Donors are drawn to have opposite number to each base sentence. $N = 32$ base examples for \textsc{Simple} and $N=192$ otherwise, balanced across sg/pl-base.}
\label{tab:cross-template-matrix}
\end{table}

We see that interventions across templates achieve comparable sufficiency to interventions within templates. The \textsc{Simple} template is the odd one out: circuit activations on other templates transfer into the \textsc{Simple} template effectively, but not vice versa.

\paragraph{Generalization to perturbations.}
We test whether the circuit is sensitive to absolute positions of the subject and prediction tokens by considering 6 perturbations of the PP template described in App.~\ref{app:synthetic-templates-extended}: adjective on subject, adjective on object, adverb insertion, clause-initial prefix, locative prefix, and longer distractor. For each perturbation, we apply interventions on $\circuit$ across pairs differing in subject number and report sufficiency. Results can be seen in \Cref{tab:gen-perturbations}. We find that the circuit is sufficient for most perturbations, staying within $81\% - 88\%$ range, compared to the PP template's $83\%$. Adding an adverb after the object is the exception, which suggests that the circuit readout components may be specific to predictions after NPs. Follow-up work could test this by holding constant the circuit components before the readout position and searching for relevant readout neurons.

\begin{table}
\centering
\small
\begin{tabular}{l c c c c}
\toprule
Perturbation                      & $\gapbase$ & $\gapintv$   & $\gapsrc$ & $\operatorname{Suff.}$ \\
\midrule
None                & $+3.48$ & $-2.30$ & $-3.48$ & $83\%$ \\
Adj on subj              & $+3.26$ & $-2.18$ & $-3.29$ & $83\%$  \\
Adj on obj               & $+3.08$ & $-2.34$ & $-3.19$ & $86\%$  \\
\textbf{Adv after obj}      & $+3.70$ & $+0.56$ & $-3.87$ & $\mathbf{42\%}$ \\
Single prefix                & $+3.00$ & $-1.88$ & $-3.00$ & $81\%$ \\
Loc. prefix                & $+4.33$ & $-3.02$ & $-4.31$ & $85\%$ \\
Long distract                 & $+3.11$ & $-2.43$ & $-3.17$ & $88\%$ \\
\bottomrule
\end{tabular}
\caption{On-circuit intervention on the perturbed \textsc{PP} template ($N = 192$). The circuit applies to all settings apart from adding an adverb after the distractor object.}
\label{tab:gen-perturbations}
\bigskip
\begin{tabular}{l c c c c}
\toprule
Setting                                   & $\gapbase$ & $\gapintv$  & $\gapsrc$ & $\operatorname{Suff.}$ \\
\midrule
Baseline & $+3.48$ & $-2.32$ & $-3.48$ & $83\%$ \\
\midrule
Held-out subjs           & $+3.69$ & $-2.26$ & $-3.58$ & $82\%$  \\
Held-out verbs        & $+2.19$ & $-1.51$ & $-2.16$ & $85\%$  \\
\bottomrule
\end{tabular}
\caption{The circuit generalizes on the \textsc{PP} template with held-out subjects or with held-out verbs. $N=192$ pairs per condition. Held-out subjects experiment samples from 13 novel subjects. Held-out verb experiment averages continuation probabilities across 36 novel verb singular/plural pairs.}
\label{tab:gen-lexical}
\end{table}

\paragraph{Generalization to novel words.}
So far, we have evaluated our circuit on a limited set of 16 subjects and objects, and measured probabilities on only six verb forms. Here, we show that the circuit is not only general across templates and perturbations, but robust to the choice of lexical items in the templates.

For this test, we create 13 single-token singular/plural subject pairs not in the original 16-pair lexicon, and 36 single-token singular/plural verb pairs. Examples of these are shown in App.~\ref{app:generalization-lexical}. 

We then intervene on $\circuit$, and find that it indeed applies to novel subjects and verbs (\Cref{tab:gen-lexical}). The lower baseline gap in held-out verbs reflects the model's weaker preference for less common verb completions, but the intervention still brings the gap $85\%$ of the way to the natural target.

\subsection{DAS analysis of the transport channel}\label{app:das}
% Supports ``H and its subspace'' in Sec.~\ref{sec:circuit}. It should contain
% DAS training details and the matched necessity/sufficiency interventions on
% the full head, the learned 1D channel, and its orthogonal complement. Move the
% existing ``DAS Details'' section here with label app:das; consider adding a
% direct main-text reference to App.~\ref{app:das}.
\paragraph{Training details.}

We use a rotation alignment map with 1D intervention. The DAS intervention is applied to $\layerhead{13}{7}$'s value-output channel at the prediction position. In \model{}, this is the 256-dimensional head slice inside the attention value output for $\layerhead{13}{7}$. Training uses the \textsc{PP} training set with batch size $4$, $5$ epochs, Adam optimizer with learning rate $5\cdot 10^{-3}$, and three random seeds.

\paragraph{Ablation evaluation.}
In \Cref{sec:circuit}, we demonstrate the sufficiency of the 1D channel for controlling agreement behavior. In this section, we further evaluate mean ablation variants on the \textsc{PP} evaluation set. We first compute the mean \layerhead{13}{7} pre-output-projection activation at the prediction position over the base examples. Let $x\in\mathbb{R}^{256}$ be the natural \layerhead{13}{7} slice for an example, let $\bar{x}$ be the mean slice, and let $v$ be the normalized 1D DAS direction.

For necessity, we compare three ablations:
\begin{align}
\text{full:}\quad
    x' &= \bar{x}, \\
\text{1D channel:}\quad
    x' &= x + (\langle \bar{x}, v\rangle - \langle x, v\rangle)v, \\
\text{complement:}\quad
    x' &= \bar{x} + (\langle x, v\rangle - \langle \bar{x}, v\rangle)v .
\end{align}
The full ablation replaces the entire 256-dimensional head slice with its mean. The 1D-channel ablation replaces only the DAS direction component with its mean value and preserves the orthogonal complement. The complement ablation does the reverse, mean ablating the orthogonal complement while preserving the natural DAS-direction component.

\end{document}